%% file: main.tex
\documentclass[11pt,a4paper]{lumia}

\usepackage[sort&compress]{natbib}
\setcitestyle{square}
\usepackage{array}
\usepackage{boldline}
\usepackage{makecell}
\usepackage{ragged2e}
\usepackage{xurl}
\usepackage{xspace}
\usepackage[section]{placeins}
\usetikzlibrary{decorations.pathreplacing,calc,arrows.meta}


\definecolor{gwaccent}{HTML}{4B3A8C}
\colorlet{lumiaabsbg}{gwaccent!7!white}
\colorlet{gwtablehead}{gwaccent!18!white}
\colorlet{gwtablerow}{black!4}
\colorlet{gwhighlight}{gwaccent!11!white}
\renewcommand{\titlefont}{\color{black}\normalfont\bfseries\fontsize{19}{23}\selectfont}

\setlist[itemize]{leftmargin=18pt, itemsep=1pt, topsep=2pt}
\setlist[enumerate]{leftmargin=20pt, itemsep=1pt, topsep=2pt}
\newcolumntype{Y}{>{\RaggedRight\arraybackslash}X}
\newcolumntype{L}[1]{>{\RaggedRight\arraybackslash}p{#1}}
\newcolumntype{C}[1]{>{\Centering\arraybackslash}p{#1}}
\colorlet{gwrule}{black!22}
\newcolumntype{I}{!{\color{gwrule}\vrule width 0.4pt}}
\newcommand{\gwtablestyle}{\scriptsize\setlength{\tabcolsep}{3.5pt}%
  \renewcommand{\arraystretch}{1.2}\hyphenpenalty=5000\exhyphenpenalty=5000}
\newcommand{\zr}{\rowcolor{gwtablerow}}
\newcommand{\hr}{\rowcolor{gwhighlight}}

\newcommand{\no}{\leavevmode{\color{black!45}$\times$}}

\setheadertext{\textcolor{gwaccent}{GameWM}}
\setheadertitle{\textcolor{gwaccent}{\textbf{GameWM}}\quad\textcolor{black!55}{Technical Report}}
\setreportdate{2026.11}

\title{\vspace{-1mm}
  \textcolor{gwaccent}{GameWM:} Toward Next-Generation Video Games\\
  with Interaction-Centric World Models
}

\author{
  \mbox{\href{https://hebing-sjtu.github.io/}{\textcolor{black}{Bing He}}$^{*}$},\enspace \mbox{\href{https://rainzor.github.io/}{\textcolor{black}{Runze Wang}}$^{*,\dagger}$},\enspace \mbox{\href{https://char-1232323.github.io/zhafei.io/}{\textcolor{black}{Fei Zha}}$^{*}$},\enspace
  \mbox{Zhongyuan Wu},\enspace \mbox{Ke Xu}\\
  \Authfont\mbox{Tao Zhuang},\enspace \mbox{Yuxuan Xie},\enspace \mbox{\href{https://digiartomnis.github.io/index.html}{\textcolor{black}{Zhongyuan Liu}}$^{\ddagger}$}
  \par\vspace{-6pt}

  {\normalfont\fontsize{8}{10}\selectfont Lightspeed Studios, Tencent}
  \par\vspace{-6pt}
}

\correspondingemail{%
  \emailicon\ \textcolor{black}{\textnormal{davidbinghe@gmail.com}}
  \quad
  \href{https://github.com/hebing-sjtu/GameWM}{\textcolor{black}{{\faGithub}\, Code}}
  \quad
  \href{https://game-world-model.github.io/}{\textcolor{black}{{\faGlobe}\, Project Page}}
  \quad
  \mbox{$^{*}$\,Equal Contribution}
  \quad
  \mbox{$^{\dagger}$\,Project Lead}
  \quad
  \mbox{$^{\ddagger}$\,Corresponding Author}
  \vspace{-1mm}
}

\makeatletter
\let\gamewm@oldmaketitle\maketitle
\renewcommand{\maketitle}{%
  \gamewm@oldmaketitle
  \par
  \begingroup
    \centering
    \begin{tikzpicture}
      \node[inner sep=0, anchor=south west] (teaser) at (0,0)
        {\includegraphics[width=\textwidth]{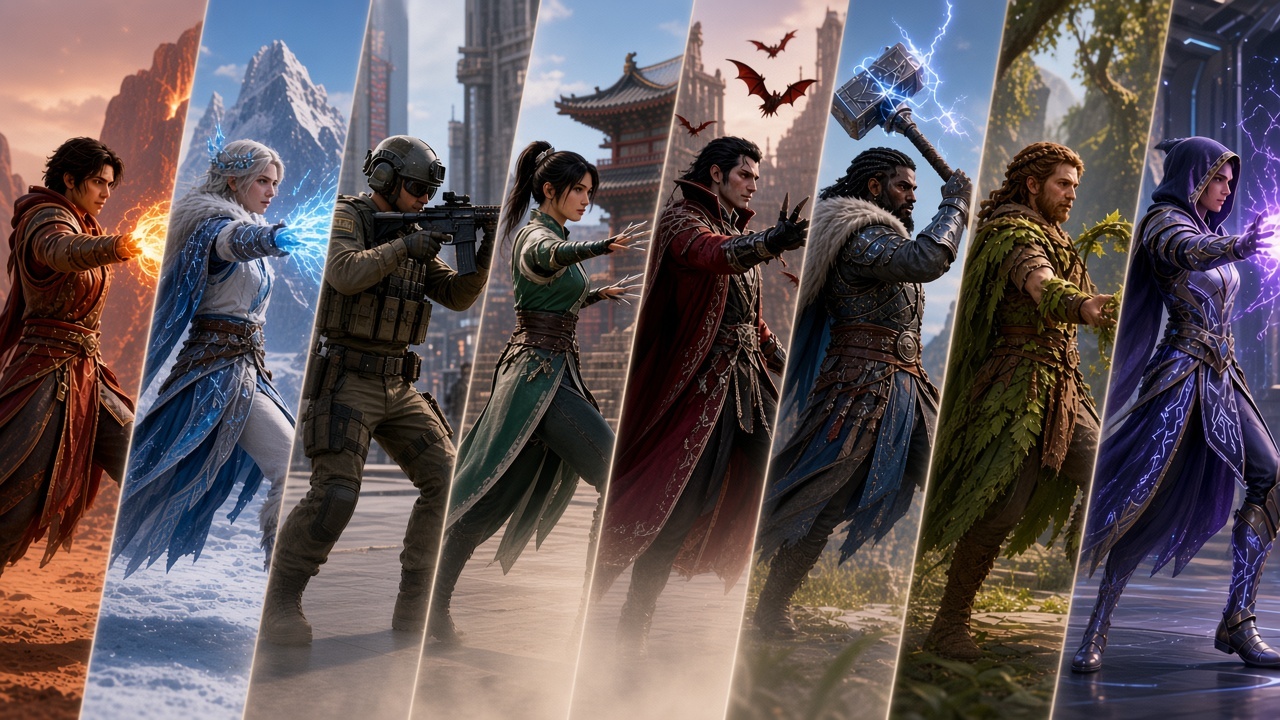}};
      \begin{scope}[x={(teaser.south east)}, y={(teaser.north west)}]
        \path[fill=white,fill opacity=.58,draw=white,draw opacity=.85,line width=.45pt]
          (0.305,0.075)--(0.685,0.075)--(0.720,0.285)--(0.340,0.285)--cycle;
        \path[fill=gwaccent,fill opacity=.035,draw=gwaccent,draw opacity=.22,line width=.35pt]
          (0.320,0.088)--(0.672,0.088)--(0.703,0.272)--(0.351,0.272)--cycle;
        \node[inner sep=0] at (0.51,0.205)
          {\fontsize{27}{27}\selectfont\bfseries\textcolor{gwaccent}{Game}\textcolor{gwaccent!68!black}{WM}};
        \node[inner sep=0] at (0.51,0.141)
          {\fontsize{7.4}{7.8}\selectfont\bfseries\color{black!82}Stable world~\textperiodcentered~Customized interaction};
        \draw[gwaccent!75!black,line width=.45pt] (0.462,0.121)--(0.558,0.121);
        \node[inner sep=0] at (0.51,0.101)
          {\fontsize{7.4}{7.8}\selectfont\bfseries\color{black!82}More than just a renderer};
      \end{scope}
    \end{tikzpicture}\par
    \captionsetup{type=figure}%
    \caption{\textbf{One mechanic, customized realizations.} The GameWM-engine defines one attack mechanic; the GameWM-VM realizes it as each character's own skill in that character's own world.}%
    \label{fig:teaser}%
    \par
  \endgroup
}
\makeatother

\hypersetup{pdftitle={GameWM: Toward Next-Generation Video Games with Interaction-Centric World Models}, pdfauthor={GameWM Team}}

\begin{document}

\input{sec/0_abstract}

\maketitle

\input{sec/1_intro}
\input{sec/2_background}
\input{sec/3_related}
\input{sec/4_design}

\input{sec/5_data}
\input{sec/6_models}
\renewcommand{\topfraction}{0.92}\renewcommand{\textfraction}{0.06}\renewcommand{\floatpagefraction}{0.85}\setcounter{topnumber}{2}\setcounter{totalnumber}{3}
\input{sec/7_experiments}
\input{sec/9_conclusion}

\newpage
\bibliographystyle{neurips_2025}
\bibliography{references}

\clearpage
\appendix
\input{sec/A_appendix}

\end{document}

%% file: sec/0_abstract.tex
\begin{abstract}
For more than half a century, video games have pursued better graphics, more natural interaction, and more engaging gameplay. Generative video models (VMs) now bear on all three. Neural networks already draw most of the pixels in shipped games, and recent world models let a lightweight game engine keep the world state while a VM turns the engine's coarse proxy video into final frames. These systems, however, treat the VM as a renderer that only restyles motion the engine has already authored, which leaves most of its generative capacity unused and limits the diversity a game can offer. We present \textbf{GameWM}, a design for next-generation games built on a clear division of labor. The GameWM-engine keeps what must be exact: rules, state, camera, and the layout of the world, which also spares the video model from having to remember it. The GameWM-VM generates what is expensive to author by hand: the look of the world and the detail of the player's interactions. One game mechanic can then be realized in many visually and physically distinct ways, while its outcome stays exactly as the engine computed it. To realize this division of labor, we build GameWM as a complete system: a data engine centered on interactions rather than exploration, an interface through which authored character sheets, scene art, and skill descriptions enter the model, and a model family from 5B to 33B parameters with a real-time causal variant for play. Together they give every player a fresh version of the same game: each player's own character, world, and skills turn the same engine mechanics into a new experience, while every event keeps the timing and outcome the GameWM-engine computed.
\end{abstract}

%% file: sec/1_intro.tex
\section{Introduction}
\label{sec:intro}

\begin{quote}
  \centering\emph{``A game is a series of interesting choices.''}\quad --- Sid Meier
\end{quote}

Video games began as laboratory curiosities, \emph{Tennis for Two} in 1958 and \emph{Spacewar!} in 1962, and became an industry with \emph{Pong} and the Magnavox Odyssey in 1972. Today 3.6B people play them~\citep{newzoo2025}. Across these seven decades, successive generations of hardware and software have pursued the same three goals: better graphics, more natural interaction, and more engaging gameplay (Section~\ref{sec:background}). Progress on one axis has repeatedly opened the others. Hardware 3D acceleration made first-person shooters and open worlds possible. Networking turned games into persistent shared worlds. Accessible creation tools let players build and share worlds of their own.

Generative video models are the newest technology to touch all three axes. \textbf{On graphics,} neural networks already produce most of the pixels in shipped games, although not yet through video generation: DLSS 4 upscales and interpolates frames so that 15 of every 16 displayed pixels are generated~\citep{nvidia2025dlss4}, DLSS 4.5 raises that to 23 of 24, and DLSS 5, released in September 2026, adds a neural rendering stage that infuses engine frames with photoreal lighting and materials~\citep{nvidia2026dlss5a,nvidia2026dlss5b}. In parallel, open video models have grown from Wan2.1 (1.3B and 14B, February 2025)~\citep{wan2025wan} to MiniMax H3 (33B, August 2026)~\citep{minimax2026h3}, and their visual quality now approaches that of film footage. The open question is therefore no longer whether generated pixels belong in games, but how much of the game a VM should be responsible for.

\textbf{On interaction,} we distinguish two levels of semantic granularity. Coarse-grained interaction moves the camera and the protagonist through the world: where to look and where to go. Fine-grained interaction is what the protagonist does to an object or another character and how: opening a car door and getting in, striking an opponent, casting a spell. 

World models have gone through two generations, and both have centered on the coarse level (Section~\ref{sec:related}). The first is driven by low-dimensional action signals, camera poses or a handful of keys~\citep{valevski2024gamengen,bruce2024genie,he2025matrixgame2,parkerholder2025genie3}, and does not yet deliver even coarse-grained control: a key says which way to go, but how fast the protagonist moves and how fast the camera turns are left to the model, so the same input yields different motion from one sample to the next. Because the world state lives only in generated frames, control degrades further under fast viewpoint changes and revisited places drift~\citep{xiao2025worldmem,ye2026mind}. A game built this way cannot be steered reliably, let alone played with immersive control. The second generation conditions the VM on a proxy video: an engine, increasingly one written by a coding agent, keeps the state and renders a coarse proxy, and the VM turns it into the final frames~\citep{chen2026cwm,zhan2026magpie,matrixgame2026v4,huang2026edenlive,nvidia2026streamrender}. Because the proxy fixes layout and camera frame by frame and is re-rendered whenever the player returns, this route makes coarse-grained interaction exact and alleviates the memory problem, and it is becoming the mainstream for playable worlds.

Fine-grained interaction, where new gameplay lies, has remained largely unexplored. Second-generation world models cast the VM as a renderer: they pair each proxy with frames that a game engine rendered or a camera recorded~\citep{zhan2026magpie,matrixgame2026v4,huang2026edenlive,chen2026cwm}, so a proxy has exactly one correct output, and none of these systems treats protagonist interaction as the objective of data curation. Appearance may be restyled, but what happens and how it moves is fixed by the pair, which leaves most of the VM's generative capacity unused. Consider a ranged attack. The engine resolves the mechanic exactly: who attacks, which target is hit at which frame, and how much damage is dealt; how the attack looks is not part of the mechanic. A VM could realize the same event as a fireball, an ice lance, or a bolt of lightning, or as a pistol shot or a rifle burst, with the attacker's motion following the weapon: a staff is raised and swung, a bow is drawn and released, a rifle is shouldered and recoils. A renderer cannot, because its data never pairs one proxy event with more than one realization. This is the increment a VM can bring to a game: not a restyled copy of the engine's frames, but fine-grained interactions that the engine never animated, played out on outcomes that the engine still computes. Because the realization follows the player's own character, world, and skills, the same mechanics unfold as different interactions for every player, and each player meets their own version of the game.

\textbf{On gameplay,} existing work is even less concrete. A position paper proposes interactive generative video as the basis of a generative game engine and outlines a five-level maturity roadmap~\citep{yu2025position}. Matrix-Game 4.0~\citep{matrixgame2026v4} lets a coding agent create gameplay from natural language, Magpie~\citep{zhan2026magpie} targets the prototyping stage of game development, and Project Genie~\citep{google2026genie} offers 60-second explorable worlds but, by Google's own account, is not a game engine~\citep{register2026genie}. These works describe what generative games could be, but not what a VM adds to a game that its engine cannot: a session is specified by a prompt, a first frame, or an engine scene, and who the protagonist is, what world it lives in, and what it can do are left unspecified.

We present \textbf{GameWM}, a system design for next-generation games in which the VM is a first-class component with a defined responsibility. Building on the engine--VM split of proxy-conditioned systems~\citep{chen2026cwm}, GameWM names the two sides the GameWM-engine and the GameWM-VM and designs every part of the system around the contract between them: a division of labor, an authoring interface through which a game supplies its characters, scenes, and skills, a data engine curated for interactions, a model family that implements the GameWM-VM for offline content and real-time play, and an evaluation that sets every output beside the proxy that drove it and the event timing. The GameWM-engine is defined by this contract rather than by an implementation: any engine, hand-written or produced by a coding agent, fills the role if it supplies what it computes exactly and cheaply, namely rules, state, camera, layout, and the coarse kinematics of the protagonist and the objects it interacts with, as a proxy video and a log of events. The GameWM-VM generates what is expensive to author: the appearance of the world and the articulated detail of protagonist interactions, and it may realize one engine event in many ways as long as the event's semantics are preserved. A game is then authored as rules in the GameWM-engine plus a character sheet, scene concept art, and skill descriptions that the GameWM-VM renders.

Our main contributions are summarized as follows:
\begin{itemize}[leftmargin=*]
  \item \textbf{A system design that assigns the VM a responsibility.} The design splits a game into a GameWM-engine, specified by the contract it must meet, and a GameWM-VM, and turns the VM from a renderer into the source of a fresh, personal experience for every player: the same mechanics, played as the player's own character, in the player's own world, with the player's own skills, while every event keeps the timing and outcome the GameWM-engine computed. This places GameWM between proxy renderers, which only restyle the motion the engine authored, and pure world models, which fail to maintain a stable world.
  \item \textbf{An interaction-centric data engine.} It mines interaction-rich clips from unpacked game buffers and from screen recordings of many titles, and turns them into aligned pairs with four proxy modalities, stylized targets, interaction-centric captions, and VLM-based quality control. Proxies of screen recordings come from monocular estimates rather than engine access, so any footage can enter and target appearance is not bounded by any one engine.
  \item \textbf{A model family from 5B to 33B that implements the GameWM-VM.} All members share one proxy-reference interface trained with LoRA or full fine-tuning, without extra control modules, and are mapped to deployment tiers: GameWM-H3-33B for offline content and as the distillation teacher, GameWM-Wan2.2-14B as a low-cost offline model, and a causal GameWM-Wan2.2-5B for real-time play; GameWM-H3-33B is also distilled into a four-step causal model that generates chunk by chunk. We evaluate them visually on held-out interaction clips, with every output shown beside the proxy that drove it and the recorded event times marked.
\end{itemize}

%% file: sec/2_background.tex
\section{Seven Decades of Video Games along Three Axes}
\label{sec:background}

\begin{figure}[!h]
  \centering
  \includegraphics[width=\textwidth]{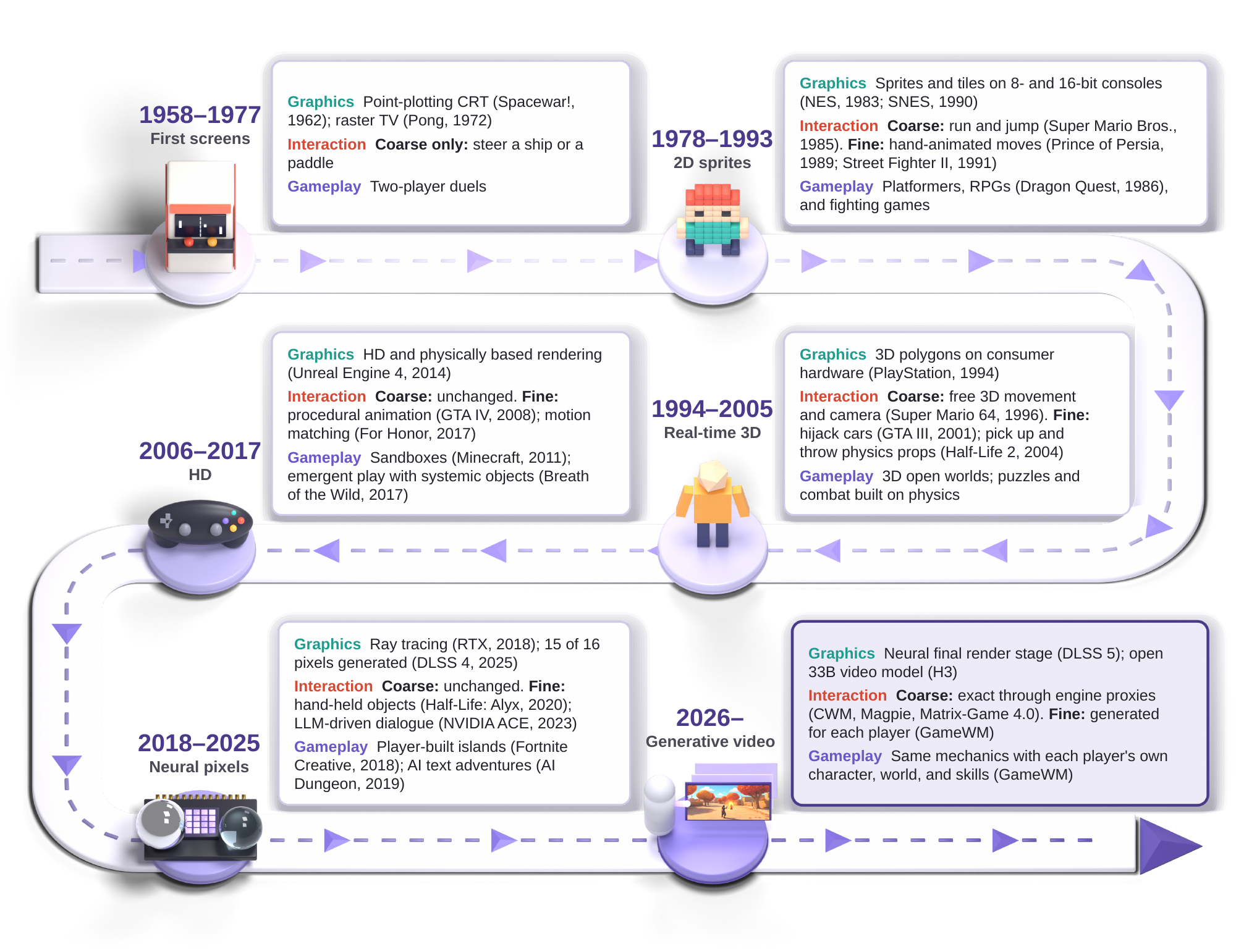}
  \caption{\textbf{Seven decades of games along graphics, interaction, and gameplay.} Eras are delimited by transitions in graphics, and each dated landmark is placed in the era of its release. Interaction is split as in Section~\ref{sec:intro}: coarse-grained interaction moves the camera and the protagonist; fine-grained interaction is what the protagonist does to objects and characters.}
  \label{fig:milestones}
\end{figure}

Across the eras of Figure~\ref{fig:milestones}, one cycle repeats in three steps, and it tells us where a VM should sit in a game.

\paragraph{A machine takes over one stage of authoring.} Real-time 3D replaced hand-drawn sprites, shaders and ray tracing compute lighting that artists once baked, physics middleware and motion matching move the objects and bodies once scripted or keyed, and DLSS reconstructs most of the pixels the engine once shaded. Coarse-grained control has needed no new takeover since the mid-1990s. Fine-grained interaction is still authored per character: every action the protagonist can perform on an object or a character, and how it looks, is designed in advance. This is the stage a VM can take over.

\paragraph{The machine joins the engine under a contract.} None of these steps replaced the engine. Each came with a contract, such as middleware interfaces, or for DLSS~5 an engine frame that defines ``what must remain''~\citep{nvidia2026dlss5b}, and the engine stayed authoritative over content and rules. Pure world models instead ask the industry to replace the engine, its rules, and its tools at once; a Google spokesperson has said that Genie ``is not a game engine and can't create a full game experience''~\citep{register2026genie}. GameWM keeps the engine and makes the contract explicit: the GameWM-engine computes every event, and the GameWM-VM decides how it plays out.

\paragraph{The new capacity becomes new gameplay.} Real-time 3D brought open worlds, physics brought puzzles and combat built on objects, and systemic objects brought the emergent play of \emph{Breath of the Wild}. DLSS makes games look better but leaves their play unchanged, and so does a VM used only as a renderer. Because fine-grained interaction is authored once per character, every player sees the same actions; a VM that takes over this stage can realize them anew for each player, as the player's own character with the player's own skills (Section~\ref{sec:authoring}).

%% file: sec/3_related.tex
\section{Related Work}
\label{sec:related}

\paragraph{Interactive video world models.} Single-game simulators~\citep{valevski2024gamengen,alonso2024diamond,decart2024oasis,kanervisto2025muse,guo2025mineworld,zhang2025matrixgame} and latent-action models learned from gameplay video~\citep{bruce2024genie} were followed by open-domain models driven by keyboard and camera~\citep{he2025matrixgame2,li2025gamecraft,mao2025yume,parkerholder2025genie3,sun2025worldplay,robbyant2026lingbot,wang2026matrixgame3,riemanndynamics2026matrixgame35,dreamxteam2026dreamx,zhu2026sanawm,alayaworld2026v11,huang2026solarwm} and by recent real-time, long-horizon systems on consumer hardware~\citep{abot2026world0,overworld2026waypoint}; open full-stack frameworks now cover the path from a bidirectional video model to a few-step autoregressive world model~\citep{zhao2026minwm,rui2026biwm,li2026forgewm}. Interaction-oriented variants widen the action vocabulary with object interactions, character actions such as combat and spell-casting, and text events~\citep{xiong2026actworld,robbyant2026lingbot2}, and ActWorld builds a dedicated interaction dataset for this setting; the results nonetheless remain coarse in both control precision and visual quality. Revisit consistency is addressed with explicit memory~\citep{xiao2025worldmem,huang2025memoryforcingspatiotemporalmemory,yu2026mosaicmem}, and memory, interaction adherence, and persistent state are measured by dedicated benchmarks~\citep{ye2026mind,ying2026wbench,lu2026wrbench}; \citet{kong2025worldmodelingsurvey} survey the wider field. These models form the first generation of interactive world models: control is limited to what a few keys can express, results stay coarse, and the state lives in the generated frames, so rules and outcomes are visual guesses (top of Table~\ref{tab:systems}).

\paragraph{Engine-conditioned world models and neural renderers.} CWM introduced a coding agent that keeps state in code, a compiler that renders a proxy video, and H3 as the renderer~\citep{chen2026cwm}. Magpie, Matrix-Game 4.0, EdenLive, and StreamRender-H3~\citep{zhan2026magpie,matrixgame2026v4,huang2026edenlive,nvidia2026streamrender} build real-time systems on the same separation within three months of 2026, from industry labs (Skywork, Mogo AI, VAST, NVIDIA) and academic groups, which makes engine- or agent-supplied proxies the emerging mainstream for playable worlds; DLSS 5 brings neural enhancement of engine frames to shipped games~\citep{nvidia2026dlss5b}. All of them train the VM to render a proxy as one fixed output; CWM leaves local motion to the VM's prior, but still learns one realization per proxy. Table~\ref{tab:systems} places their designs between the first generation and GameWM. On the state side, agents and programs increasingly maintain the world that a neural model renders~\citep{liao2026streamingcwm,huang2026programmable,liang2026codevideomodel,chi2026agentgarten}.

\begin{table}[t]
  \centering
  \caption{\textbf{Interactive world models compared by design choice.} Entries come from each paper or project page; inputs, resolutions, and hardware differ, so the table compares design choices rather than measured performance. ``Fine-grained interaction'': what the user can do beyond moving and looking, and who produces it. n/r: not reported in the sources we read.}
  \label{tab:systems}
  \gwtablestyle\fontsize{6.5}{7.6}\selectfont\setlength{\tabcolsep}{2.4pt}\renewcommand{\arraystretch}{1.08}
  \begin{tabularx}{\textwidth}{L{2.1cm}IC{0.5cm}IYL{1.7cm}YL{1.72cm}C{1.45cm}Y}
    \hlineB{2.5}
    \rowcolor{gwtablehead}
    \textbf{System} & \textbf{Year} & \textbf{Control / proxy} & \textbf{Backbone} & \textbf{Paired data} & \textbf{Real time} & \textbf{VM decides} & \textbf{Fine-grained interaction} \\
    \hlineB{1.5}
    \multicolumn{8}{l}{\textit{First generation: action-conditioned, state in the generated frames}} \\
    \hline
    GameNGen & 2024 & DOOM key actions & Stable Diffusion v1.4, augmented & RL-agent play of DOOM & 20 FPS, 1 TPU & Everything & The game's own actions \\
    \zr Oasis & 2024 & Keyboard & DiT + ViT autoencoder, 500M open & Minecraft & 20 FPS & Everything & Mine, place, pick up, open chests \\
    Hunyuan-GameCraft & 2025 & Keys + mouse as camera motion & HunyuanVideo & ${>}$1M recordings, 100+ AAA games; synthetic & 6.6 FPS, distilled & Everything & Navigation only \\
    \zr Matrix-Game 2.0 & 2025 & Frame-level keys + mouse & SkyReels-V2-1.3B (Wan2.1) & UE and game capture (${\sim}$1,200\,h); Sekai & 25 FPS & Everything & Navigation only \\
    Genie 3 & 2025 & Navigation; text world events & n/r & n/r & 720p, 24 FPS & Everything & Text events (weather, objects, characters), not by the agent \\
    \zr WorldPlay & 2025 & Keys + mouse (dual action) & n/r & 320K real and synthetic clips, 170K from AAA games & 720p, 24 FPS, 8~GPUs & Everything & Navigation only \\
    ActWorld & 2026 & Keys + mouse; interaction commands & Wan2.1-14B & 100K synthesized interaction clips; 400\,h real walking & n/r & Everything & Object interaction, 40 action categories \\
    \zr LingBot-World 2.0 & 2026 & Actions; text events & 14B; 1.3B light & Egocentric, game and UE, web video & 720p, 60 FPS, multi-GPU & Everything & Attack, archery, spells, shooting; text events \\
    ABot-World-0 & 2026 & 8 keys: movement + camera & n/r & AAA games, simulation engines, web video & 720p, 16 FPS & Everything & Navigation only \\
    \hlineB{1.5}
    \multicolumn{8}{l}{\textit{Engine-conditioned: state in an engine or agent, the VM renders a proxy}} \\
    \hline
    CWM & 2026 & Depth + semantic ID & H3-33B & GTA~V gameplay (5.6\,h); KITTI-360 proof of concept & \no & Appearance & Engine; local motion from VM prior, one realization \\
    \zr Magpie & 2026 & White-box & Wan2.2-5B & Human-played UE scenes (300\,h) & 32 FPS; ${\sim}$1.55\,s input to response & Appearance & Engine-animated; events logged, not used in training \\
    Matrix-Game 4.0 & 2026 & Low-poly proxy; dynamic captions & Wan2.2-5B & Dual-pass UE5 capture (1,712\,h) & 480p, 60 FPS & Appearance & Engine-animated \\
    \zr EdenLive & 2026 & Any reference (DINOv3) & Wan2.2-5B; Wan2.1-14B & Real and game video; agent-built proxies & ${>}$24 FPS & Appearance & Engine-animated \\
    StreamRender-H3 & 2026 & Semantic map (13 classes) & H3-33B & Engine sessions with AI drivers & 2-step, 8~GPUs & Appearance & Engine-animated \\
    \hline
    \hr \textbf{GameWM} & 2026 & DUV, depth, semantics, low-poly & Wan2.2-5B; Wan2.2-14B; H3-33B & Game buffers; screen recordings; stylized targets & 4-step causal students at 5B/14B/33B; 33B first chunk 3.1\,s on 8~GPUs & \textbf{Appearance + interaction} & \textbf{Generated by the GameWM-VM: several variants per event, chosen by text; curated data} \\
    \hlineB{2.5}
  \end{tabularx}
\end{table}

\paragraph{Spatially controlled video generation.} ControlNet-style and reference-based control~\citep{zhang2023controlnet,tan2024ominicontrol,jiang2025vace,ju2025fulldit,bian2026videoasprompt,nvidia2025cosmostransfer,nvidia2025cosmos}, 3D- and motion-aware conditioning~\citep{gu2025das,burgert2025gowiththeflow}, and photorealism enhancement of synthetic or G-buffer renders~\citep{richter2023epe,liang2025diffusionrenderer,wang2025zeroshotrealism} condition a video model on depth, edges, poses, motion, or G-buffers. Concurrent generative renderers start from G-buffers~\citep{huang2026worldrenderer,zeng2026rgbx}, coarse 3D simulations or lightweight proxies~\citep{gomeznogales2026coarsetoreal,xu2026proxy2world}, or structured states exported from physics engines~\citep{alaya2026renderer}, and SolarRenderer follows arbitrary control videos through H3's reference pathway with frame-aligned positions~\citep{fang2026solarrenderer}. These methods treat the control signal as a complete description to be preserved, most of them through an added control branch. GameWM's proxy need not carry the protagonist's limb motion or the effects of an interaction, which the model adds, and it enters through the backbone's own reference pathway as a prefix with its own temporal coordinates (Section~\ref{sec:models}).

\paragraph{Causal generation and few-step distillation.} Real-time world models generate chunk by chunk with an autoregressive video diffusion model. Diffusion forcing~\citep{chen2024diffusion} trains causal generation with per-frame noise levels, and CausVid~\citep{yin2025slow} distills a bidirectional teacher into a causal student; Self Forcing~\citep{huang2025selfforcing} trains the student on its own rollouts to close the gap between clean training history and generated inference history, and later work refines this recipe and bounds error accumulation over long rollouts~\citep{zhu2026causal,zhao2026causalforcingscalablefewstep,zhao2026maskforcing,zhuang2026selfgradientforcing,huang2026livelonghorizoninteractivevideo}. Distribution matching~\citep{yin2024one,yin2024dmd2} and flow-map distillation~\citep{gu2026anyflowanystepvideodiffusion} reduce sampling to a few steps. GameWM follows the SolarWM pipeline~\citep{huang2026solarwm} built from these components and conditions both teacher and student on the proxy.

\paragraph{Data for world models.} Datasets for world modeling pair video with actions in games~\citep{guss2019minerl,yu2025gamefactory,li2026wildworld}, with camera trajectories and spatial annotations in real and rendered footage~\citep{li2025sekai,he2026sekai2,wang2025spatialvid,zhou2025omniworld}, or with simulator buffers~\citep{dosovitskiy2017carla,zhong2024unrealzoo}, and SolarWM unifies ten such sources into one frame-aligned format with metric camera geometry~\citep{huang2026solarwm}. They are organized around exploration, camera motion, or action labels; GameWM's corpus is instead cut around protagonist interactions, and each proxy is paired with several targets rather than one (Section~\ref{sec:data}).

%% file: sec/4_design.tex
\section{GameWM System Design}
\label{sec:design}

GameWM is designed for one experience: an immersive game in which the player's own character, weapons, and skills look the same and change the world in the same way in every clip, scene, and session, while every rule and outcome stays exact. This section derives the system from that goal. Section~\ref{sec:spectrum} places GameWM between engine-bound graphics and pure world models, Section~\ref{sec:division} divides each frame between the GameWM-engine and the GameWM-VM, and Section~\ref{sec:authoring} shows how a game is authored once so that what the GameWM-VM generates stays consistent. The models that take these assets as input, and the deployment tiers they serve, follow in Section~\ref{sec:models}.

\subsection{Where GameWM Sits: Stability and Generative Capability}
\label{sec:spectrum}

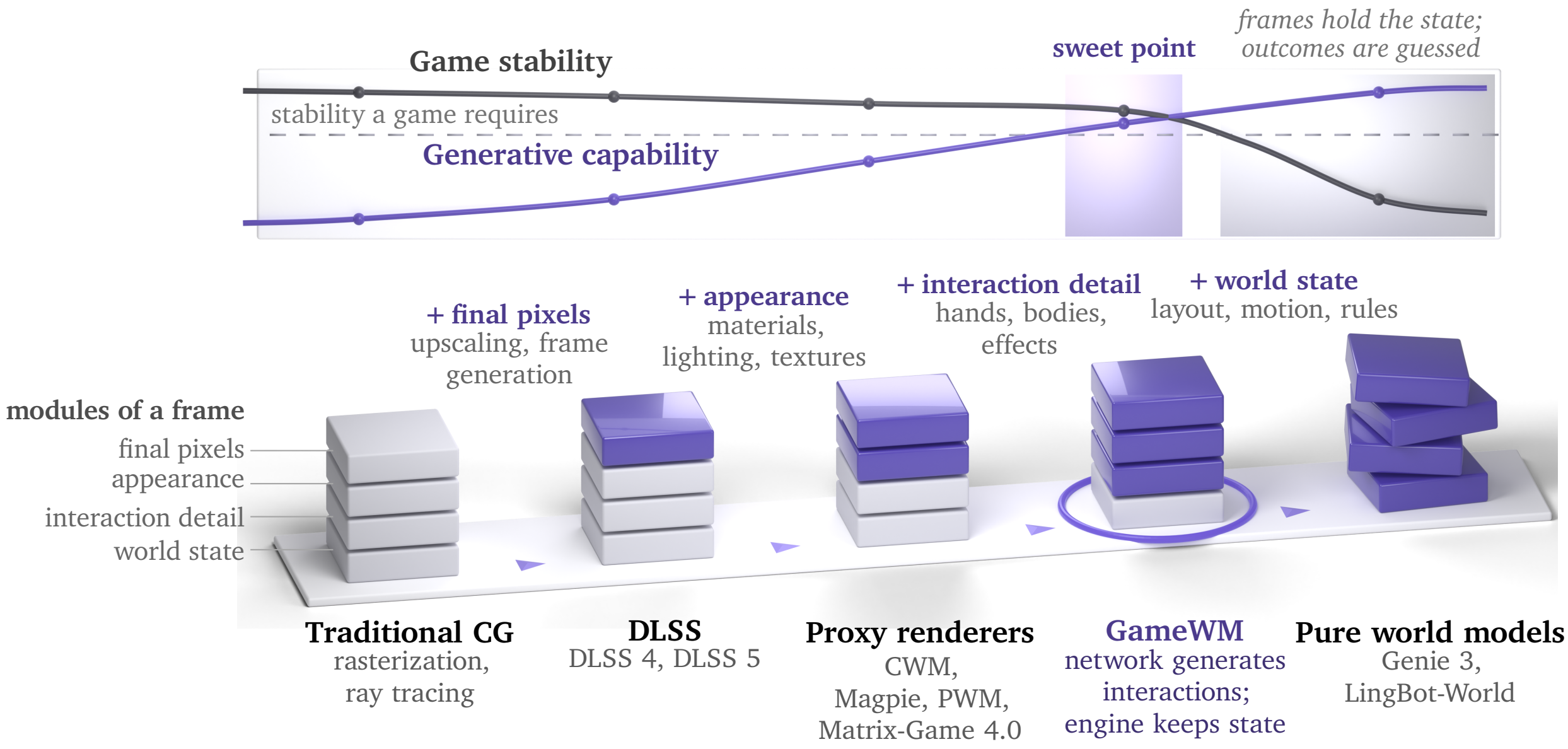
\begin{figure}[t]
  \centering
  \input{figs/fig_spectrum3d.tex}
  \caption{\textbf{GameWM maximizes generative capability subject to game stability.} Designs run from traditional CG (left) to pure world models (right). Each is a stack of the four modules of a frame, with world state at the bottom: purple blocks are produced by a neural network, grey blocks by the engine, and the label above each gap names the module the network takes over between two neighboring designs. The curves are schematic, not measured.}
  \label{fig:authority}
\end{figure}

    From traditional graphics to a pure world model, each design hands one more module of the frame to a neural network (Figure~\ref{fig:authority}). DLSS takes over the final pixels and refines what the engine rendered~\citep{nvidia2025dlss4,nvidia2026dlss5b}. Proxy renderers, the engine-conditioned world models such as CWM, Magpie, PWM, and Matrix-Game 4.0, take over appearance and render it from a coarse proxy video rather than from finished engine frames~\citep{chen2026cwm,zhan2026magpie,huang2026programmable,matrixgame2026v4}. GameWM also hands over interaction detail: how each interaction plays out. Pure world models such as Genie 3 and LingBot-World hand over the rest, namely layout, motion, rules, and state~\citep{parkerholder2025genie3,robbyant2026lingbot}. Each step adds generative capability, the range of content the network produces without being authored, and gives up some game stability, the guarantee that rules, state, and outcomes are exact and replayable. Stability declines slowly while the engine keeps the world state, because a visual error cannot change an outcome; it collapses once the state moves into the generated frames and outcomes become visual guesses. GameWM takes the last step before that collapse: it maximizes generative capability subject to game stability, with the network deciding how interactions look and the engine still deciding where everything is and what happens.

\subsection{Division of Labor}
\label{sec:division}

The GameWM-engine keeps everything a designer must be able to inspect, replay, and balance: rules, scores, health, collisions, quest state, and the positions and coarse poses of all entities. It renders a proxy video from the player's camera. The proxy determines where things are and when an interaction happens, for example that the protagonist reaches the driver's door at second 2.1 and the car departs at second 3.4. The GameWM-VM determines how the interaction looks: the hand on the handle, the body folding into the seat, the door closing, and the appearance of the protagonist, the car, and the street. A visual error can therefore make the feedback less clear, but it cannot change an outcome, because outcomes are computed before the GameWM-VM runs.

This division follows from cost. The quantities on the GameWM-engine side are cheap to compute exactly and expensive for a VM to infer: a rule is a line of code, and a revisit is a re-render. The quantities on the GameWM-VM side are expensive to author and cheap to sample from a strong prior: a new interaction costs an animator weeks, but a VM trained on interaction footage has seen thousands of examples of people entering cars.

\paragraph{One event, many realizations.} Because outcomes live in the GameWM-engine, the GameWM-VM is free to vary everything that does not change them. An engine event fixes its semantics, such as the event type, the participants, and the hit frame, together with the coarse kinematics of the bodies involved; any realization that keeps these is valid, so one ranged attack can appear as a fire spell, an ice spell, or a rifle shot. The proxy therefore has to carry event semantics rather than a fully posed animation, and the data has to pair one event type with many realizations (Section~\ref{sec:data-curation}). Section~\ref{sec:authoring} uses this freedom to give every character its own skills.

\subsection{Authored Once: Consistency for an Immersive Game}
\label{sec:authoring}

\begin{figure}[!t]
  \centering
  \includegraphics[width=\textwidth]{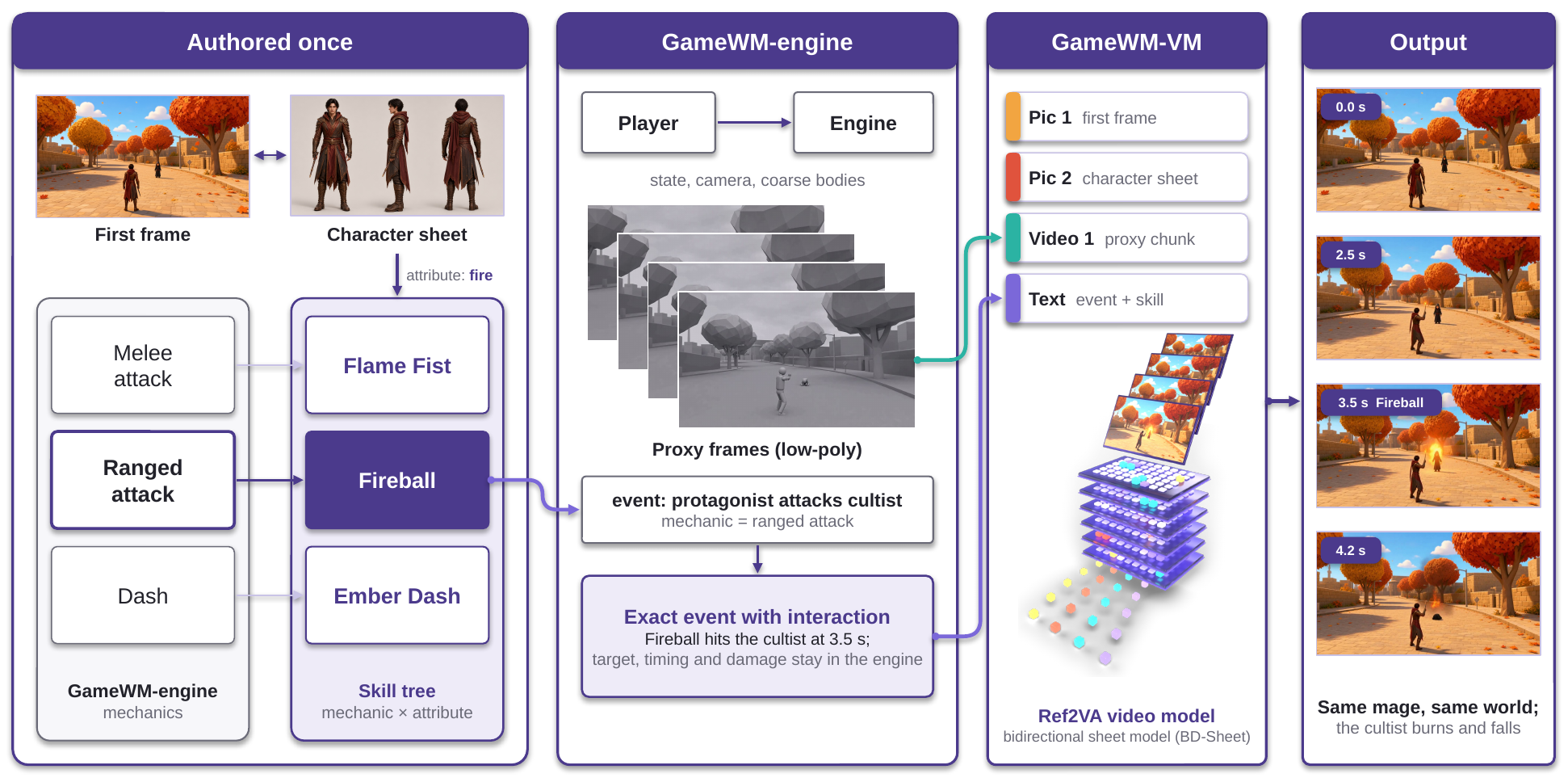}
  \caption{\textbf{From authored assets to play.} Left: authored once, a first frame and a character sheet, and a skill tree that fuses an attribute of the sheet (fire) with each mechanic. Middle: at run time the GameWM-engine renders low-poly proxy frames and resolves the event, here a ranged attack that becomes Fireball. Right: the four inputs fill the slots of the GameWM-VM's reference interface (H3's Ref2VA, Section~\ref{sec:models}); the output column shows frames of a BD-Sheet sample generated from them.}
  \label{fig:infer-pipeline}
\end{figure}

What GameWM offers the player is immersion, and immersion rests on consistency. A player who plays a fire mage expects the same face and costume in every scene, the same fire in every attack, and the same scorch where that fire lands; if any of these changes from one clip to the next, the game stops feeling like one world. Game world models so far set up each session from a prompt, a first frame, or an engine scene, and nothing in the system carries the protagonist's identity or abilities from one session to the next. GameWM makes them assets that are authored once and reused by every clip: the mechanics of the GameWM-engine, the character and its world, and the skills that join the two (Figure~\ref{fig:infer-pipeline}).

\paragraph{Mechanics from the GameWM-engine.} The game framework defines a small set of mechanics, for example a melee attack, a ranged attack, and a dash. Each is code with exact rules: what can be targeted, when the hit lands, and how much damage it does. The designer writes each mechanic once, every character shares it, and it is what the proxy shows: coarse bodies in motion and an event in the engine's log.

\paragraph{The character and its world.} A character sheet shows the protagonist from the front, the side, and the back. In conventional production a character is a textured, rigged 3D model with an animation set. Under the division of labor this asset splits in two: the GameWM-engine keeps a coarse body, a capsule or low-poly mesh that it simulates, collides, and renders into the proxy, and the sheet carries everything the player sees, at the cost of an image instead of modeling, rigging, and animation. The first frame, scene concept art aligned with the GameWM-engine's opening view, sets the look of the world. The two are generated to agree with each other, starting from whichever the player provides. From a sheet, an image model paints the first frame from that opening view with the character in it; this is the step that restyles first frames in the data engine, so the anchor relates to the proxy at inference as it did in training. From a first frame, the image model draws a sheet of its protagonist. The sheet slot is trained with exact supervision, because each stylized target in the data engine was rendered from its sheet (Section~\ref{sec:data-targets}), and the sheet is dropped for a fifth of those samples, so the first frame alone can also set both the world and the protagonist.

\paragraph{Skills: mechanic $\times$ attribute.} A character also implies how it fights. High-level attributes are read from the sheet, such as an element, a weapon, and a palette, and each attribute is fused with each engine mechanic into one entry of the character's skill tree. For the fire mage, the melee attack becomes Flame Fist, the ranged attack Fireball, and the dash Ember Dash; an ice mage on the same mechanics gets frost versions of all three. Each entry is text that describes the skill's look, including what it does to the scene, such as the burst on impact or the embers left on the ground. A designer thus writes each mechanic once and obtains a family of skills across characters, and a player can wield skills that no artist animated.

\paragraph{From assets to frames.} At run time the GameWM-engine turns the player's input into proxy frames and resolves each event exactly. When an event uses a mechanic, the character's entry for that mechanic is added to the caption of the chunk in which it occurs, and the first frame, the sheet, the proxy chunk, and the caption enter the GameWM-VM together as in-context references (Section~\ref{sec:interface}): motion, framing, and positions follow the proxy, the world's look follows the first frame, the protagonist's identity follows the sheet, and the caption describes the event. The GameWM-engine still decides the target, the hit frame, and the damage; the entry decides only how the skill looks, and the GameWM-VM generates the attacker's motion, the projectile, the impact, and the target's reaction in step with the hit frame.

Together these assets make the experience consistent across a game rather than within a clip: the same protagonist appears in every scene, each scene opens with its authored look, every use of a skill draws on the same description, and its timing and outcome come from the same engine rule. Figure~\ref{fig:exp-oc-scenes} places four characters in four scenes each, and Figure~\ref{fig:exp-oc-mechanics} gives eight characters their own skills on a ranged and a melee mechanic (Section~\ref{sec:exp}).

%% file: figs/fig_spectrum3d.tex
\begin{tikzpicture}[nohyph/.style={execute at begin node={\hyphenpenalty=10000\exhyphenpenalty=10000}},dname/.style={nohyph, font=\footnotesize\bfseries, align=center, text width=2.9cm, inner sep=1pt},desc/.style={nohyph, font=\scriptsize, align=center, text width=2.45cm, inner sep=1pt, black!65},trans/.style={nohyph, font=\scriptsize, align=center, text width=2.6cm, inner sep=1pt, anchor=south},lay/.style={font=\scriptsize, black!60, anchor=east, inner sep=1.5pt}]
  \node[anchor=south west, inner sep=0] (img) at (2.25cm,0) {\includegraphics[width=\dimexpr\linewidth-2.75cm\relax]{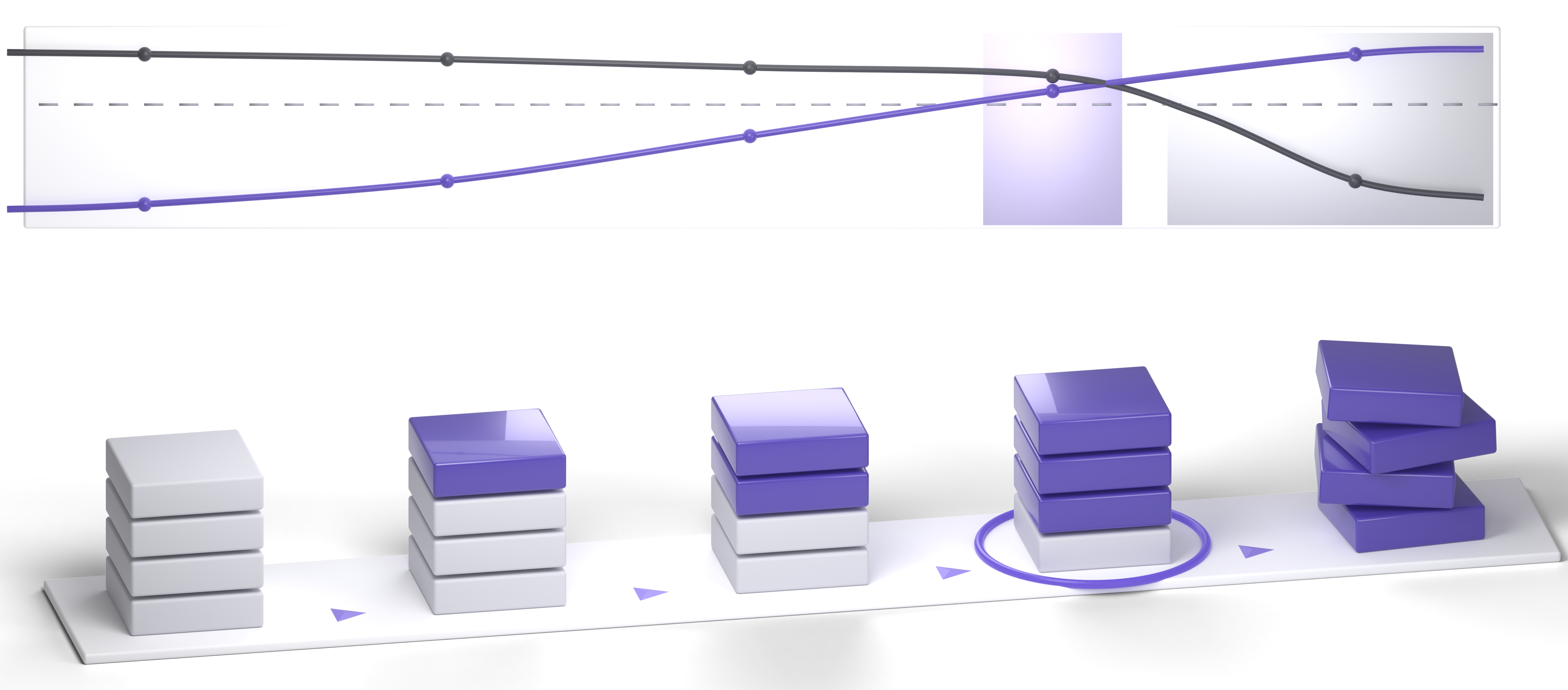}};
  \begin{scope}[x={($(img.south east)-(img.south west)$)}, y={($(img.north west)-(img.south west)$)}, shift={(img.south west)}]
    \coordinate (l0) at (0.0753,0.1363);
    \node[lay] (lt0) at ([xshift=4pt]img.west |- l0) {world state};
    \draw[black!35, line width=0.4pt] (lt0.east) -- ([xshift=-1pt]l0);
    \coordinate (l1) at (0.0753,0.1930);
    \node[lay] (lt1) at ([xshift=4pt]img.west |- l1) {interaction detail};
    \draw[black!35, line width=0.4pt] (lt1.east) -- ([xshift=-1pt]l1);
    \coordinate (l2) at (0.0753,0.2497);
    \node[lay] (lt2) at ([xshift=4pt]img.west |- l2) {appearance};
    \draw[black!35, line width=0.4pt] (lt2.east) -- ([xshift=-1pt]l2);
    \coordinate (l3) at (0.0753,0.3064);
    \node[lay] (lt3) at ([xshift=4pt]img.west |- l3) {final pixels};
    \draw[black!35, line width=0.4pt] (lt3.east) -- ([xshift=-1pt]l3);
    \node[font=\scriptsize\bfseries, black!70, anchor=south east, inner sep=1.5pt] at ([yshift=2pt]lt3.north east) {modules of a frame};
    \node[trans] at (0.2060,0.4047) {\textcolor{gwaccent}{\textbf{+\,final pixels}}\\[-1pt]{\color{black!60}upscaling, frame generation}};
    \node[trans] at (0.3990,0.4352) {\textcolor{gwaccent}{\textbf{+\,appearance}}\\[-1pt]{\color{black!60}materials, lighting, textures}};
    \node[trans] at (0.5921,0.4658) {\textcolor{gwaccent}{\textbf{+\,interaction detail}}\\[-1pt]{\color{black!60}hands, bodies, effects}};
    \node[trans] at (0.7851,0.5171) {\textcolor{gwaccent}{\textbf{+\,world state}}\\[-1pt]{\color{black!60}layout, motion, rules}};
    \node[dname, anchor=north] (n0) at (0.1309,0.0230) {Traditional CG};
    \node[desc, anchor=north] at (n0.south) {rasterization, ray tracing};
    \node[dname, anchor=north] (n1) at (0.3239,0.0230) {DLSS};
    \node[desc, anchor=north] at (n1.south) {DLSS 4, DLSS 5};
    \node[dname, anchor=north] (n2) at (0.5169,0.0230) {Proxy renderers};
    \node[desc, anchor=north] at (n2.south) {CWM, Magpie, PWM, Matrix-Game 4.0};
    \node[dname, text=gwaccent, anchor=north] (n3) at (0.7100,0.0230) {GameWM};
    \node[desc, text=gwaccent!80!black, anchor=north] at (n3.south) {network generates interactions; engine keeps state};
    \node[dname, anchor=north] (n4) at (0.9030,0.0230) {Pure world models};
    \node[desc, anchor=north] at (n4.south) {Genie 3, LingBot-World};
    \node[font=\footnotesize\bfseries, black!75, anchor=south west, inner sep=1pt] at (0.1273,0.9389) {Game stability};
    \node[font=\footnotesize\bfseries, gwaccent, anchor=south east, inner sep=1pt] at (0.3672,0.7794) {Generative capability};
    \node[font=\scriptsize, black!55, anchor=south west, inner sep=1pt] at (0.0228,0.8519) {stability a game requires};
    \node[font=\scriptsize\bfseries, gwaccent, anchor=south, inner sep=1pt] at (0.6713,0.9659) {sweet point};
    \node[nohyph, font=\scriptsize\itshape, black!55, anchor=south, align=center, inner sep=1pt] at (0.8507,0.9659) {frames hold the state;\\[-1pt]outcomes are guessed};
  \end{scope}
\end{tikzpicture}

%% file: sec/5_data.tex
\section{Interaction-Centric Data Engine}
\label{sec:data}

\begin{figure}[!t]
  \centering
  \includegraphics[width=\textwidth]{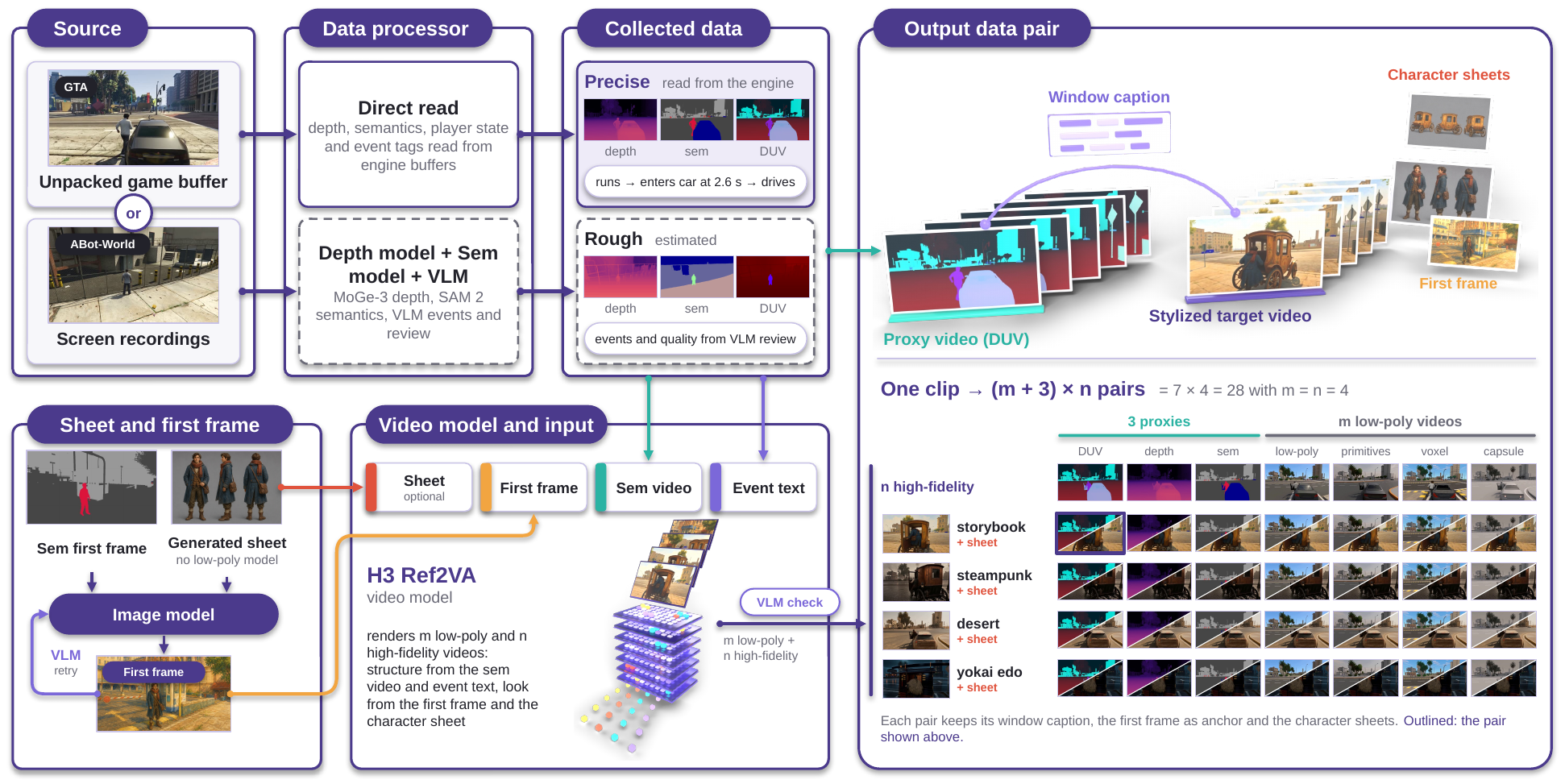}
  \caption{\textbf{The interaction-centric data engine.} Left: both source paths yield the same three proxies and an event caption, from which H3 Ref2VA renders $m$ low-poly and $n$ high-fidelity versions of the clip. After VLM screening, each of the three proxies and the $m$ low-poly videos is paired with every high-fidelity target and with the source recording. Right: one ``enter vehicle'' clip with $m=n=4$, giving $7\times4=28$ stylized pairs, plus seven with the recording (not drawn); the outlined cell is the pair shown above the grid.}
  \label{fig:data-pipeline}
\end{figure}

Figure~\ref{fig:data-pipeline} summarizes the data engine. A clip comes from one of two sources: a game recording whose depth, semantics, player state, and event tags are read from engine buffers, or a screen recording of another title, whose depth and semantics are estimated by monocular models~\citep{kong2026moge3,ravi2025sam2} and whose events are tagged by a VLM. Clips are cut around interactions and reduced to three proxy modalities (DUV, depth, and semantics) and a window-scoped event caption. H3's reference-to-video-and-audio mode (Ref2VA) then renders each clip again from its semantic video and event text: $m$ low-poly versions, which stand in for the proxy frames a GameWM-engine would render, such as a simple engine or one written by a coding agent, and serve as further proxies, and $n$ high-fidelity versions, which take their look from a restyled first frame and an optional character sheet and serve, with the source recording, as targets.

\subsection{Sources and the Appearance Ceiling}
\label{sec:data-sources}

\begin{figure}[t]
  \centering
  \includegraphics[width=\textwidth]{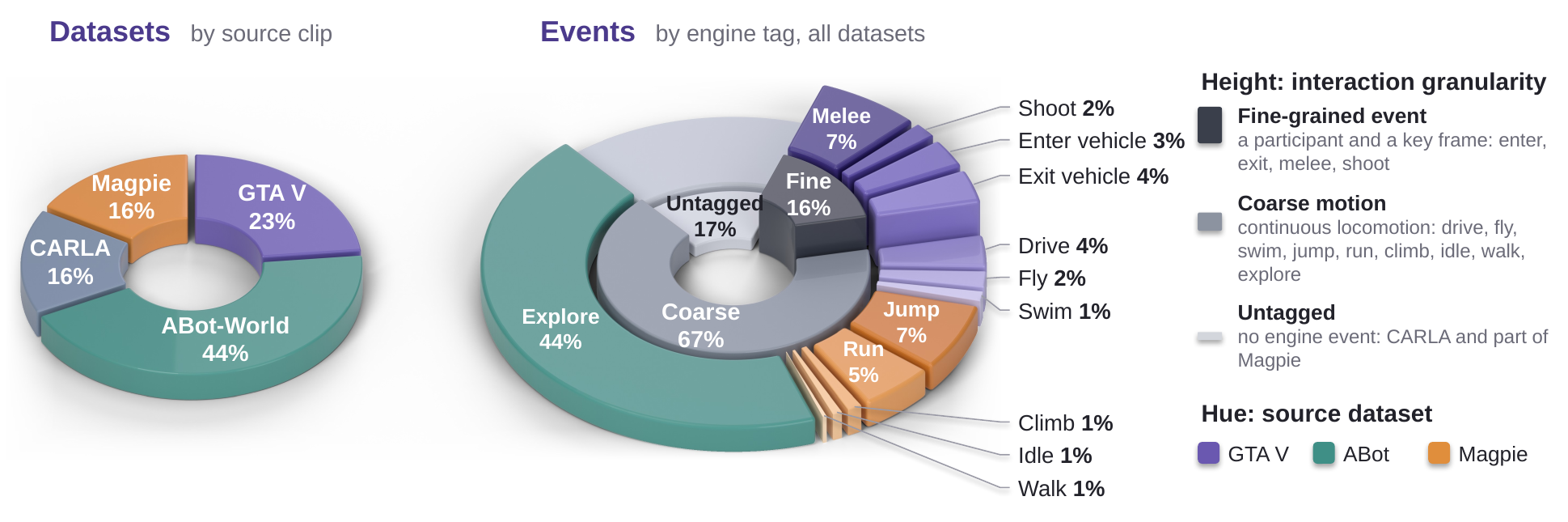}
  \caption{\textbf{Composition of the data engine by source clips.} Left: the share of each source. Right: the event of every clip, with granularity on the inner ring and the event on the outer ring; hue gives the source, and height the granularity (fine-grained, coarse, untagged). ABot-World is the released 500-hour, action-annotated part of the training data of ABot-World-0~\citep{abot2026world0}, with depth and semantics estimated by MoGe-3~\citep{kong2026moge3} and SAM~2~\citep{ravi2025sam2}; CARLA~\citep{dosovitskiy2017carla} and Magpie~\citep{zhan2026magpie} come from SolarRenderer-Data~\citep{fang2026solarrenderer}. Shares are of the snapshot in Table~\ref{tab:data-snapshot}.}
  \label{fig:data-composition}
\end{figure}

Paired data captured inside an engine has an appearance ceiling: the target is itself an engine render, so a model trained on it learns to look like that engine. GameWM therefore does not tie its data to one engine. Any footage from which per-frame depth and semantics can be obtained, either read from engine buffers or estimated by monocular models, enters the same pipeline and is reduced to the same proxy interface: per-frame depth and semantics, also packed into a single RGB video (DUV), and low-poly re-renders in four styles (amateur low-poly, capsules, Blender primitives, and MagicaVoxel), which show the structure of a scene without the engine's appearance. Appendix~\ref{sec:app-proxy} gives the encoding. Adding a source requires only an extractor; the training interface does not change. Figure~\ref{fig:data-composition} shows the sources currently converted and the events their clips carry, divided by granularity.

\subsection{Fine-Grained Curation and Coarse Expansion}
\label{sec:data-curation}

The sources play two roles, and they are curated differently.

\begin{figure}[t]
  \centering
  \includegraphics[width=0.497\textwidth]{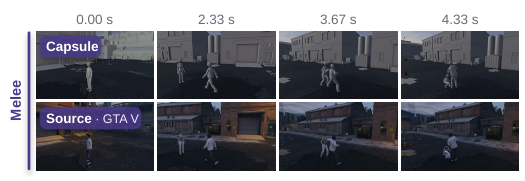}\hfill
  \includegraphics[width=0.497\textwidth]{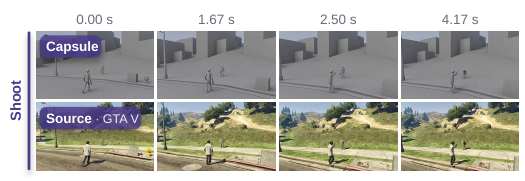}\\[1pt]
  \includegraphics[width=0.497\textwidth]{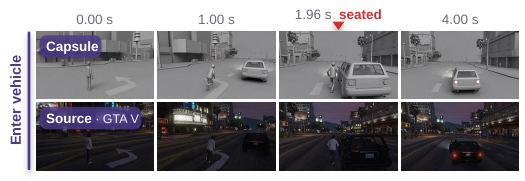}\hfill
  \includegraphics[width=0.497\textwidth]{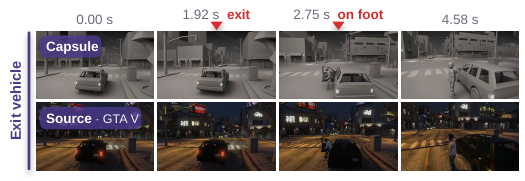}\\[1pt]
  \includegraphics[width=0.497\textwidth]{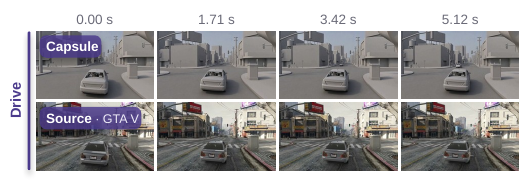}\hfill
  \includegraphics[width=0.497\textwidth]{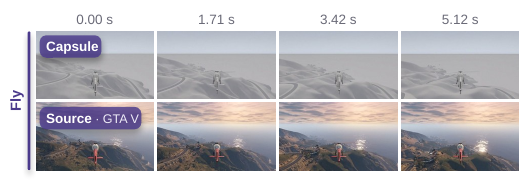}\\[1pt]
  \includegraphics[width=0.497\textwidth]{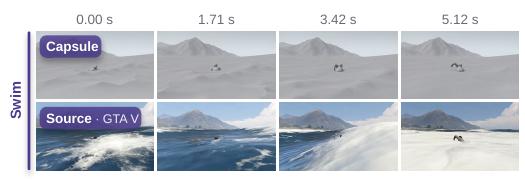}\hfill
  \includegraphics[width=0.497\textwidth]{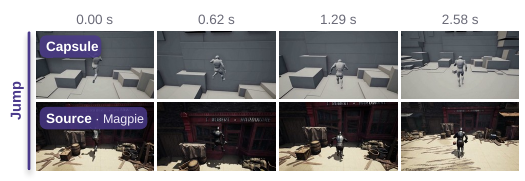}\\[1pt]
  \includegraphics[width=0.497\textwidth]{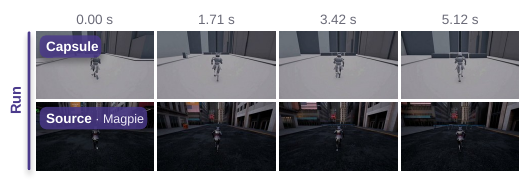}\hfill
  \includegraphics[width=0.497\textwidth]{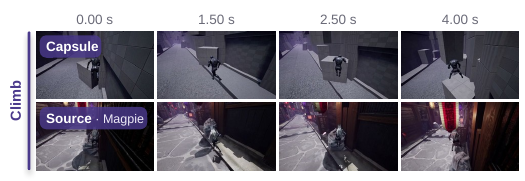}\\[1pt]
  \includegraphics[width=0.497\textwidth]{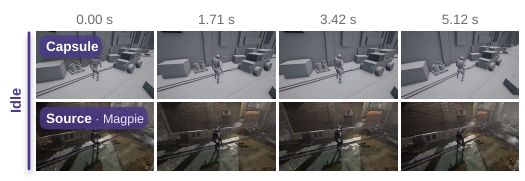}\hfill
  \includegraphics[width=0.497\textwidth]{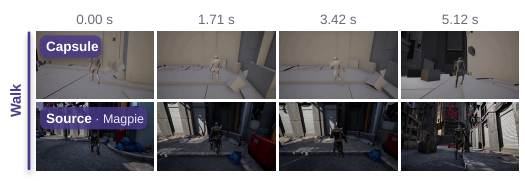}\\[1pt]
  \includegraphics[width=0.497\textwidth]{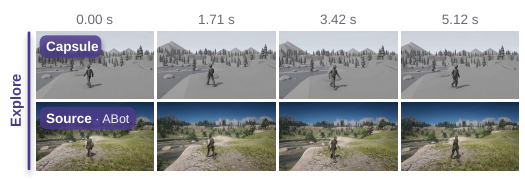}\hfill
  \includegraphics[width=0.497\textwidth]{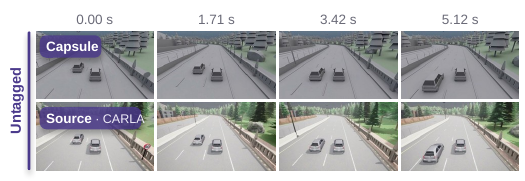}
  \caption{\textbf{One training pair per engine event.} Each block shows a clip's capsule re-render, which is the proxy, above the source recording. The first two rows are fine-grained GTA~V events; the rest are coarse or untagged clips from GTA~V, Magpie, ABot-World, and CARLA. Red ticks mark engine state changes. The capsules carry the camera, the layout, and the timing of each event, but no limbs, faces, or effects.}
  \label{fig:data-event-pairs}
\end{figure}

\paragraph{Fine-grained events teach interactions.} Where the source reports what happens, clips are cut around events rather than sampled uniformly. In the GTA~V recordings the engine reports each event, such as \nolinkurl{enter_vehicle} or \nolinkurl{shoot}, with its frame, and the event times enter the proxy and the caption. Fine-grained events have a participant and a key frame that decides the outcome: entering or exiting a vehicle, a melee hit, a shot. They are what the GameWM-VM cannot infer from motion alone and what the GameWM-engine's clock must time. Coarse events are locomotion modes, such as driving, swimming, or climbing, which the proxy already describes almost completely. Of the 1,460 GTA~V clips in the snapshot, 1,018 (70\%) are fine-grained (Figure~\ref{fig:data-composition}, right); Figure~\ref{fig:data-event-pairs} shows one training pair per event.

\paragraph{Coarse and untagged clips broaden appearance.} ABot-World, Magpie, and CARLA tag only locomotion or nothing, so their clips cannot be cut around events, but they reduce to the same proxy interface and are rendered into the same target styles. With the coarse GTA~V clips they make up 84\% of the corpus and supply far more scenes, layouts, and camera paths than one game offers, from which the model learns to take structure from the proxy and appearance from the anchor (Section~\ref{sec:authoring}). Interactions are thus learned from the fine-grained part, and appearance transfer from all of it.

One event can also look many ways. Captions name the realization (weapon, element, vehicle class) so that text can select it, but recorded realizations are still few: melee is bare-handed, and the vehicles are cars and a helicopter. Others, such as spell elements and melee weapons, are supplied at inference by the character sheet and the skill text (Figure~\ref{fig:exp-oc-mechanics}).

\subsection{Diverse Targets and Verified Captions}
\label{sec:data-targets}
\label{sec:data-captions}

A proxy describes structure; the target and the caption decide what the model learns to put on top of it. A clip's targets are its source recording and its high-fidelity renders. Each render takes the semantic proxy, a first frame restyled by an image model, and, for many renditions, a character sheet showing the protagonist or the driven vehicle from the front, the side, and the back. The render's subject is drawn from that sheet, so the sheet is an exact identity reference for it and is kept with the pair; the source recording has none. Pairing one proxy with targets of different looks forces the model to take structure from the proxy and appearance from the anchor and the sheet.

Each 5.17\,s training window gets its own caption with timestamp prefixes. Because captioners describe objects that a scene usually contains, the per-second content of the proxy is given to the captioner as a prior and used to check its reply: a caption that mentions a car in a second without vehicle pixels fails (Appendix~\ref{sec:app-data}).

\subsection{Quality Control Follows Provenance}
\label{sec:data-qc}
\label{sec:data-corpus}

Proxies are checked only where they are estimated. Depth and semantics read from engine buffers or from a simulator (GTA~V, CARLA, Magpie) are exact; those estimated from screen recordings can flicker or mislabel over time, so these clips are reviewed by a VLM for the temporal stability and accuracy of depth and semantics. Every rendered video, low-poly or high-fidelity, is generated and is therefore screened: a VLM compares it with its source clip on camera, motion and timing, spatial relations, style, and, with a sheet, identity, and a human reviewer can override any decision. Acceptance tracks the proxy (Table~\ref{tab:data-snapshot}): 94\% of the GTA~V low-poly renders and 70\% of its high-fidelity renders pass, against about half or less for renders built on estimated (ABot-World) or sparse gray-box (Magpie) proxies. Exact proxies thus pay twice: they need no screening themselves, and the targets built on them are more reliable. Appendix~\ref{sec:app-data} gives the review panels, the axes, the decision rules, and the rates per source.

%% file: sec/6_models.tex
\section{The GameWM Model Family}
\label{sec:models}

\subsection{Deployment Tiers}
\label{sec:tiers}

A model's scale decides where it can run, and where it runs decides what it can serve. GameWM is therefore a family of three models, one per deployment tier, rather than one model.

The offline tier, GameWM-H3-33B, has the highest quality. It renders cutscenes, trailers, player replays, and user-generated videos, produces training data, and serves as the teacher for distillation. A single H3 stream currently takes several datacenter GPUs, so this tier is not for live play. The player tier, a causal GameWM-Wan2.2-5B, is distilled to few-step chunk-wise generation for real-time play. Matrix-Game 4.0 shows that a 5B renderer can reach 60 FPS at 480p on one RTX 5090~\citep{matrixgame2026v4}, so this is the tier that can follow DLSS onto consumer hardware.

GameWM-Wan2.2-14B sits between the two as a low-cost offline model. It serves offline content that does not need the quality of the 33B model at a fraction of its cost, and it is the platform on which data and design choices are screened before they are run at 33B.

\subsection{Interface and Backbones}
\label{sec:interface}

All GameWM models share one interface. The inputs enter through the backbone's existing reference pathway rather than a new control branch: the proxy video as \texttt{<Video 1>}, the first frame as \texttt{<Picture 1>}, the optional character sheet as \texttt{<Picture 2>}, and a text that states each input's role and describes the event. One model accepts four proxy modalities, distinguished by a modality tag in the text: packed depth and semantics (DUV), depth, semantics, and low-poly re-renders. Backbones are adapted with LoRA~\citep{hu2021lora} or full fine-tuning only. Proxy frames are placed either as a sequential prefix or at the temporal coordinates (in the rotary position embedding~\citep{su2021roformer}) of the target frames they describe; Appendix~\ref{sec:eval-ablations} compares the two.

The models differ in backbone and in how the proxy enters it. GameWM-H3-33B places the proxy in H3's native Ref2VA reference rows and trains a rank-128 LoRA on the attention projections; the audio rows are kept but receive zero loss. GameWM-Wan2.2-14B is built on the high-noise expert of Wan2.2-I2V-A14B, which runs every denoising step (the low-noise expert is not used), and trains a rank-128 LoRA on the attention projections at $832\times480$ and 81 frames at 16\,fps. GameWM-Wan2.2-5B is built on Wan2.2-TI2V-5B and is fully fine-tuned at $1280\times704$ and 121 frames at 24\,fps. Wan2.2 has no reference slot, so the inputs of the shared interface are laid out as one token sequence for its self-attention: the pictures (first frame, then the sheet), the proxy, and the target, each patchified on its own spatial grid. The proxy is encoded at half the target's spatial resolution and placed as a temporal prefix, i.e.\ its latent frames take the time coordinates after the pictures and before the target. Table~\ref{tab:models} summarizes the three models.

\begin{table}[!b]
  \centering
  \caption{\textbf{The GameWM models.} All three share the interface of Section~\ref{sec:interface}, are trained on the same corpus, and are distilled to four-step causal students by the stages of Section~\ref{sec:models-causal}.}
  \label{tab:models}
  \gwtablestyle
  \begin{tabular}{L{3.1cm}IL{3.1cm}L{2.6cm}L{2.2cm}C{2.6cm}}
    \hlineB{2.5}
    \rowcolor{gwtablehead}
    \textbf{Model} & \textbf{Backbone} & \textbf{Proxy entry} & \textbf{Adaptation} & \textbf{Clip} \\
    \hlineB{1.5}
    \textbf{GameWM-H3-33B} & MiniMax-H3 (33B) & Native Ref2VA rows & LoRA, rank 128 & $1280\times704$, 124 frames, 24\,fps \\
    \zr GameWM-Wan2.2-14B & Wan2.2-I2V-A14B, high-noise expert (14B) & Prefix, half resolution & LoRA, rank 128 & $832\times480$, 81 frames, 16\,fps \\
    GameWM-Wan2.2-5B & Wan2.2-TI2V-5B & Prefix, half resolution & Full fine-tuning & $1280\times704$, 121 frames, 24\,fps \\
    \hlineB{2.5}
  \end{tabular}
\end{table}

\subsection{Long Rollout and Real-Time Distillation}
\label{sec:models-causal}

Long sessions are generated in windows: the first window receives one latent frame of context, and later windows receive ten. CWM uses a similar 34-frame overlap, so we only report the setting. All three models are distilled in the three stages of SolarWM~\citep{huang2026solarwm}: bidirectional adaptation, teacher-forced AnyFlow causal initialization, and DMD with self-gradient forcing, which yields a four-step causal student per backbone. In the causal stages each chunk reads only its own proxy chunk and the five before it (Eq.~\eqref{eq:causal}). Section~\ref{sec:exp-dmd} compares the three students on one clip.

%% file: sec/7_experiments.tex
\section{Experiments}
\label{sec:exp}

We examine two stages of GameWM-H3-33B on held-out interaction clips: the bidirectional model, which generates a whole clip at once, and the distilled causal model, which generates it chunk by chunk in four denoising steps. The causal students of the two Wan2.2 backbones are compared with it at the end of Section~\ref{sec:exp-dmd}. Both receive only a proxy, the first frame, a caption, and, for the Sheet variants, a character sheet. The proxies stand in for the output of a GameWM-engine: they are derived from held-out recordings, and their event times come from the recorded game's engine log or are read off the proxy. Evaluation is visual: every figure shows the proxy above the output it drives, with these event times marked, so that adherence and timing can be judged directly (Appendix~\ref{sec:app-protocol}). Because proxy-conditioned systems differ in proxy format, interface, and availability, and text prompts cannot be aligned fairly across them, we do not rank GameWM against other systems and show only our own models.

\subsection{Setup}
\label{sec:exp-setup}

\paragraph{Inputs.} Each sample is a 5.17\,s clip (124 frames at 24\,fps) rendered at $1280\times704$. The condition $\mathbf{c} = (y, \mathbf{a}, \mathbf{s}, \mathbf{p})$ consists of the caption $y$ with a modality tag and time-window prefixes, the appearance anchor $\mathbf{a}$ (the first frame), the character sheet $\mathbf{s}$ (Sheet variants only), and the proxy video $\mathbf{p}$ in one of four modalities at $640\times352$. Held-out clips are split by source recording (Appendix~\ref{sec:app-protocol}).

\paragraph{Bidirectional stage.} Let $\mathbf{z}_0$ be the target latent and $\mathbf{z}_t$ its noisy version at time $t$. The proxy enters H3's native reference rows, and a rank-128 LoRA on the attention projections is trained with the backbone's flow-matching objective~\citep{lipman2022flow}
\begin{equation}
  \mathcal{L}_{\mathrm{BD}}
  = \mathbb{E}_{\mathbf{z}_0,t,\boldsymbol{\epsilon}}
    \big\| f_{\theta}(\mathbf{z}_t, t, \mathbf{c}) - \mathbf{u}_t \big\|_2^2,
  \label{eq:bd}
\end{equation}
where $\mathbf{u}_t$ is the native velocity target. The loss covers only the video rows; the audio rows of H3 are kept and receive zero loss.

\paragraph{Causal stage.} The latent sequence is split into chunks $\mathbf{z}^{k}$ of five latent frames. The causal model factorizes the clip as
\begin{equation}
  p_{\psi}\big(\mathbf{z}^{1:K} \mid \mathbf{c}\big)
  = \prod_{k=1}^{K} p_{\psi}\big(\mathbf{z}^{k} \mid \mathbf{z}^{<k},\, \mathbf{p}^{\,k-W+1:k},\, y,\, \mathbf{a}\big),
  \qquad W = 6,
  \label{eq:causal}
\end{equation}
so chunk $k$ reads its own proxy chunk and the five before it, never a later one. This is what the GameWM-engine can supply during play: the proxy of the current moment, not of the future. Following SolarWM~\citep{huang2026solarwm}, the causal model is initialized with teacher-forced AnyFlow~\citep{gu2026anyflowanystepvideodiffusion} and then trained by distribution matching~\citep{yin2024one,yin2025slow} on its own rollouts~\citep{huang2025selfforcing}, with self-gradient forcing~\citep{zhuang2026selfgradientforcing} for gradients through the key--value cache:
\begin{equation}
  \nabla_{\psi}\mathcal{L}_{\mathrm{DMD}}
  = \mathbb{E}\Big[ J_{G_{\psi}}^{\top}\, w(t) \big( s_{\mathrm{fake}}(\mathbf{z}_t, t, \mathbf{c}) - s_{\mathrm{real}}(\mathbf{z}_t, t, \mathbf{c}) \big) \Big],
  \label{eq:dmd}
\end{equation}
where $G_{\psi}$ is the causal generator, $s_{\mathrm{real}}$ is the score of the frozen bidirectional model of Eq.~\eqref{eq:bd}, and $s_{\mathrm{fake}}$ is a critic trained on the student's rollouts. The proxy therefore conditions both the teacher and the student, so distillation matches proxy-conditioned distributions rather than unconditional video statistics.

\paragraph{Checkpoints.} Table~\ref{tab:checkpoints} lists the GameWM-H3-33B checkpoints of this section. The two AR models are listed because they initialize the DMD students; their own few-step outputs are not shown.

\begin{table}[!tbp]
  \centering
  \caption{\textbf{Checkpoints of GameWM-H3-33B in the experiments.} ``Sheet'' variants add a character sheet as a second anchor image. Sampling steps are per clip for BD and per chunk for the causal models.}
  \label{tab:checkpoints}
  \gwtablestyle
  \begin{tabularx}{\textwidth}{L{1.5cm}IL{5.5cm}YC{0.9cm}C{2.1cm}}
    \hlineB{2.5}
    \rowcolor{gwtablehead}
    \textbf{Model} & \textbf{Initialized from} & \textbf{Training} & \textbf{Steps} & \textbf{Sampling} \\
    \hlineB{1.5}
    BD & H3 & Eq.~\eqref{eq:bd}, lr $2{\times}10^{-4}$, batch 8 & 900 & 30, full clip \\
    \zr AR & BD & Teacher-forced AnyFlow, lr $10^{-4}$, batch 8 & 600 & few-step, causal \\
    \hr \textbf{DMD} & AR; teacher and critic from BD & Eq.~\eqref{eq:dmd}, 5 critic updates per student update & 1,000 & 4, causal \\
    \hline
    BD-Sheet & BD & Eq.~\eqref{eq:bd} with character sheet & +600 & 30, full clip \\
    \zr AR-Sheet & BD-Sheet & Teacher-forced AnyFlow & 900 & few-step, causal \\
    \hr \textbf{DMD-Sheet} & AR-Sheet; teacher and critic from BD-Sheet & Eq.~\eqref{eq:dmd} & 1,000 & 4, causal \\
    \hlineB{2.5}
  \end{tabularx}
\end{table}

\subsection{Bidirectional Model}
\label{sec:exp-bd}

\begin{figure}[!tbp]
  \centering
  \includegraphics[width=\textwidth]{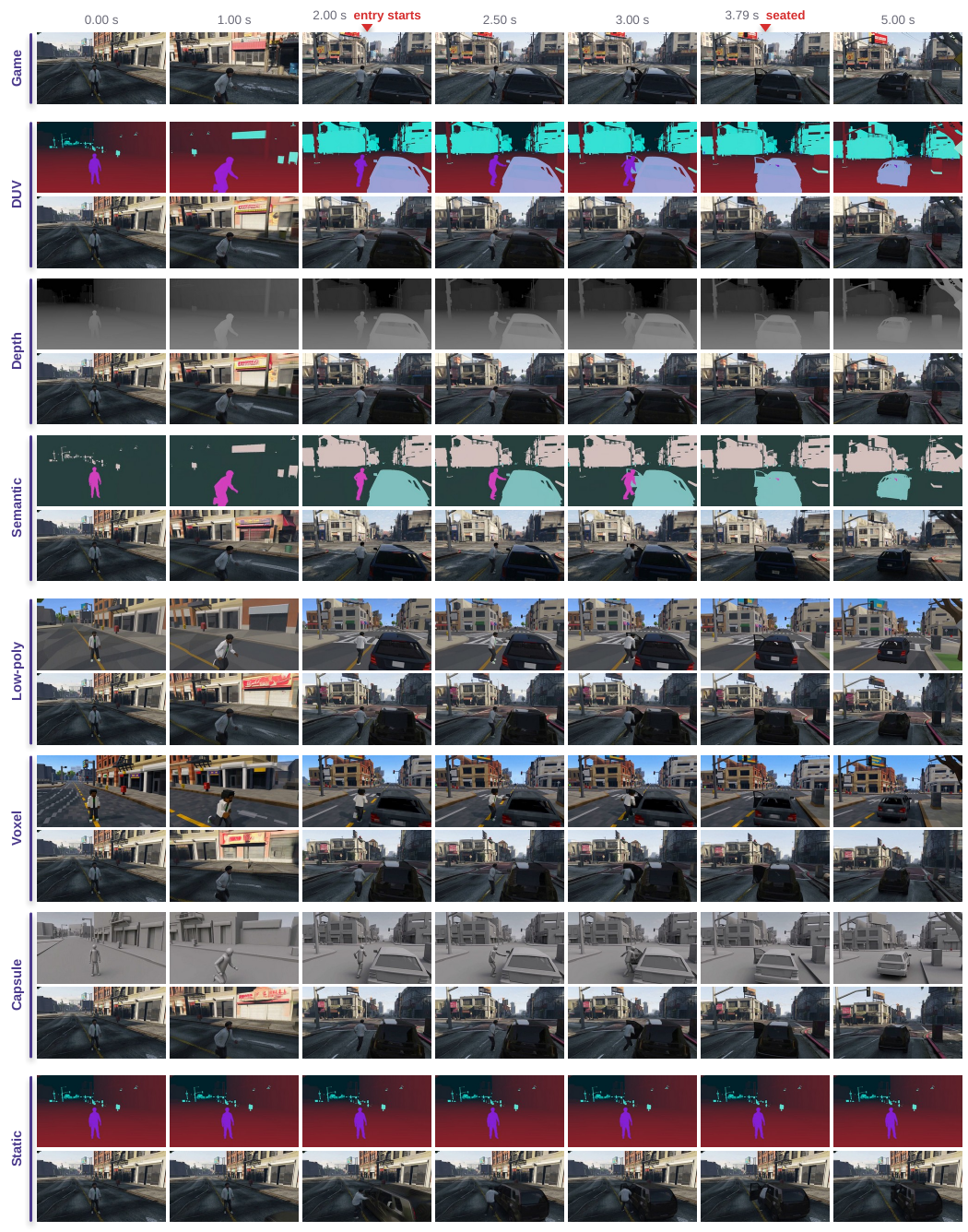}
  \caption{\textbf{Every proxy on one held-out vehicle entry.} Top: the recorded game. Below, each group shows a proxy above the bidirectional output it drives: three engine buffers (DUV, depth, semantic), three coarse re-renders (low-poly, voxel, capsule), and a static control in which the first DUV frame is held for the whole clip. Red ticks mark the engine's entry start and seated frames.}
  \label{fig:exp-granularity}
\end{figure}

\paragraph{One interaction, every proxy.}
\label{sec:eval-interaction}
Figure~\ref{fig:exp-granularity} drives one held-out vehicle entry with six proxies and a static control. Under the three engine buffers the camera, the street, and the parked cars follow the proxy, and the protagonist reaches the door and is seated at the engine's frames. The three coarse re-renders carry no limb motion, and under the capsules the protagonist is a rigid body, yet in all three he runs to the car, the door opens, and the car pulls away on the engine's schedule: the model decides how the interaction plays out (Figure~\ref{fig:authority}). Under the static control the model still stages an entry, but the camera keeps its heading and the car arrives where the model chooses, so the model reads the proxy rather than continuing the first frame.

\begin{figure}[!tbp]
  \centering
  \includegraphics[width=\textwidth]{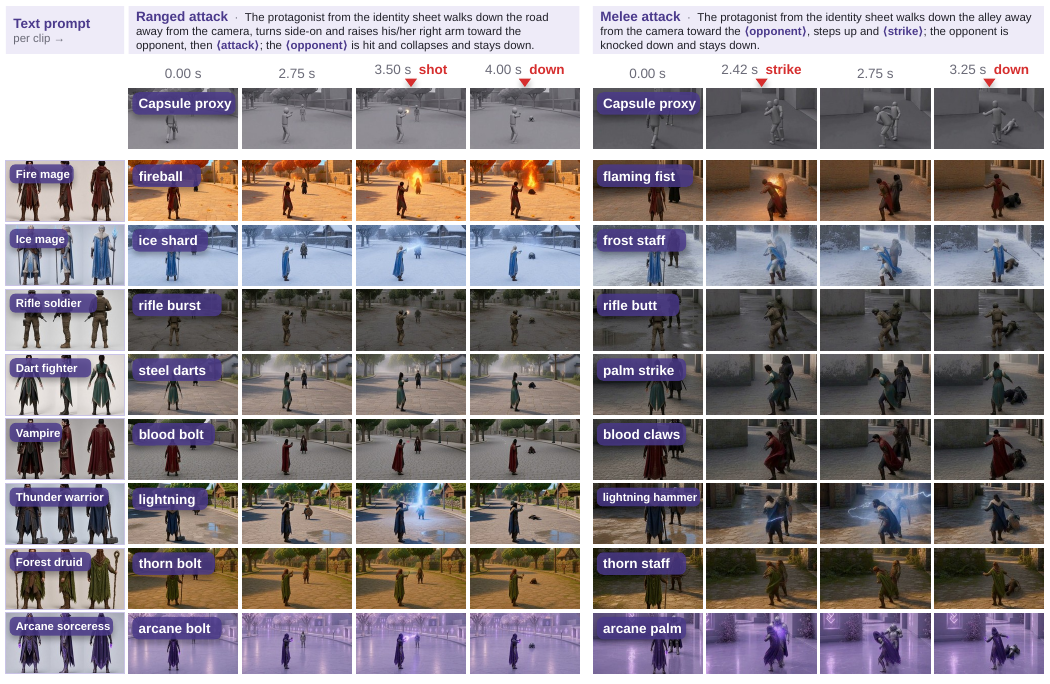}
  \caption{\textbf{One mechanic, many realizations: ranged and melee.} The left four columns are the shooting clip and the right four a melee clip; each clip keeps one capsule proxy and one caption template (the box above its columns). Each row is one character: its sheet and name (far left), a first frame of that character in each scene, and its skill for each mechanic (the tag on the first frame of each clip). BD-Sheet, 30 steps; red ticks mark the shot, the strike, and the fall as read off the proxies. The DMD-Sheet version of the ranged clip is in Figure~\ref{fig:app-oc-skills}.}
  \label{fig:exp-oc-mechanics}
\end{figure}

\paragraph{One mechanic, many realizations.} Figure~\ref{fig:exp-oc-mechanics} gives eight characters their own sheet, first frame, and skill on two capsule proxies: a shooting clip, reused in later figures, in which the protagonist attacks an opponent from a distance, and a melee clip with a single strike. In every row the opponent falls on the proxy's schedule, while everything the proxy lacks comes from the model: the capsule has no arms, so every casting pose is generated, and the melee proxy throws a bare elbow, so the flames, staffs, claws, and hammer are too. Each character keeps its costume, colors, and build in both scenes. One mechanic thus becomes eight characters' skills with nothing changed on the GameWM-engine side.

\begin{figure}[!tbp]
  \centering
  \includegraphics[width=\textwidth,height=0.87\textheight,keepaspectratio]{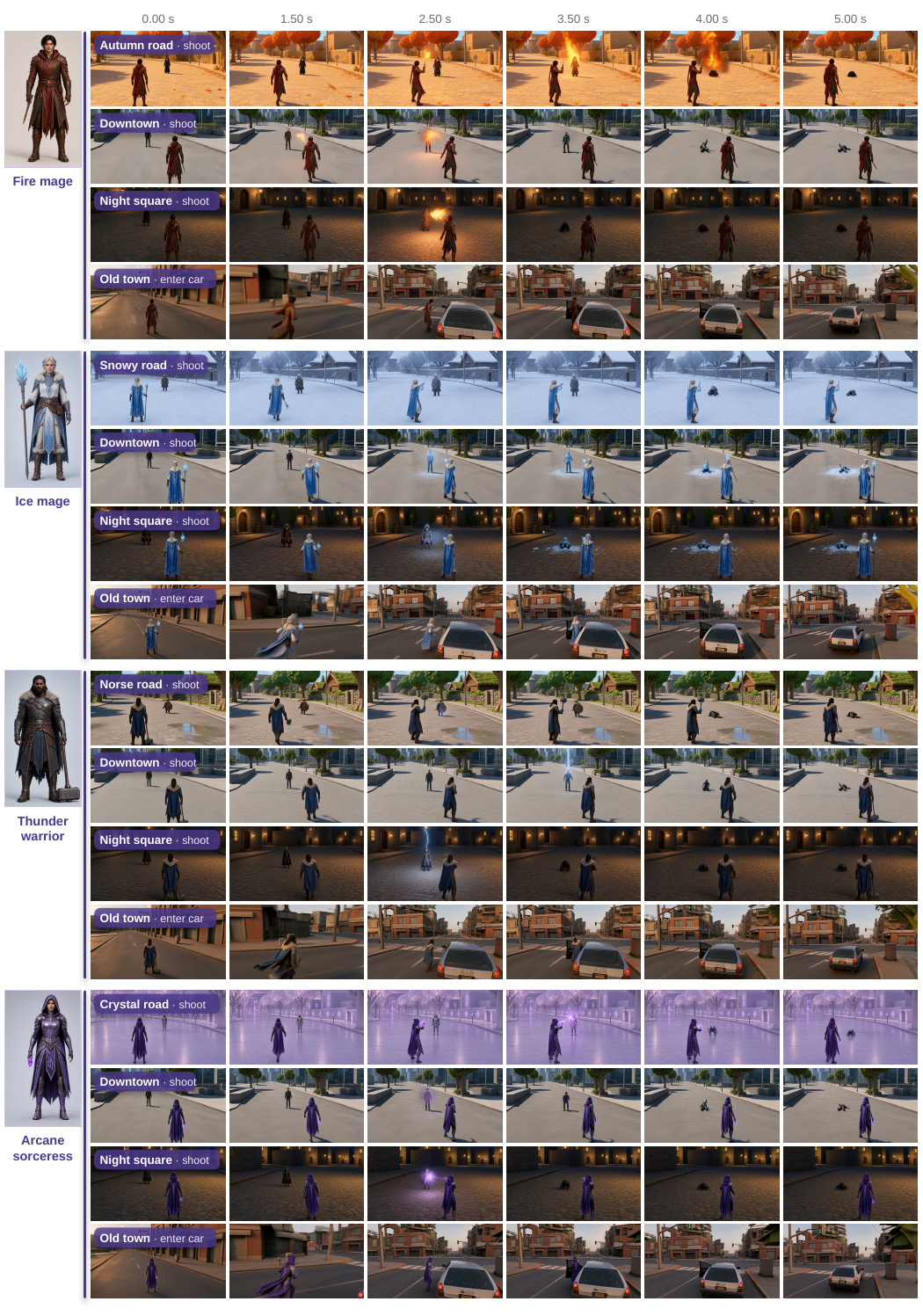}
  \caption{\textbf{One character across scenes.} Four characters (left, with their sheets), each in four scenes: three shooting clips and a held-out vehicle entry, all driven by capsule proxies. Each first frame is authored for its scene from the sheet. BD-Sheet model; columns from 0 to 5\,s. The DMD-Sheet version is in Figure~\ref{fig:app-oc-scenes-dmd}.}
  \label{fig:exp-oc-scenes}
\end{figure}

\paragraph{One character across scenes.} Figure~\ref{fig:exp-oc-scenes} places four characters in four scenes each, with first frames made from the sheet. The protagonist keeps the costume, colors, and silhouette of its sheet in all sixteen pairs, while the scene, the opponent, and the interaction follow each clip. Measured by DINOv2 similarity to the sheets (Table~\ref{tab:exp-identity}; Appendix~\ref{sec:app-protocol}), the protagonist is closer to its own sheet than to the other three in every pair, by 0.03 to 0.20. Removing the sheet but keeping the first frame lowers similarity by only 0.024 for BD-Sheet and 0.014 for DMD-Sheet, less than the 0.039 between two seeds, so in these clips identity is carried mainly by the first frame; isolating the sheet needs clips in which the protagonist enters later.

\begin{table}[!tbp]
  \centering
  \caption{\textbf{Identity across scenes.} DINOv2 cosine similarity between the tracked protagonist and a character sheet, averaged over the sixteen scene--character pairs of Figure~\ref{fig:exp-oc-scenes} (mean $\pm$ standard deviation over pairs). ``Own'': the protagonist's own sheet. ``Others'': mean over the other three characters' sheets. ``Margin'': own minus others. ``Sheet removed'': the same model, first frame, and caption, run without the sheet; similarity is still measured against the sheets.}
  \label{tab:exp-identity}
  \gwtablestyle
  \begin{tabular}{L{1.7cm}IC{1.75cm}C{1.0cm}C{1.0cm}IC{1.75cm}C{1.0cm}C{1.0cm}}
    \hlineB{2.5}
    \rowcolor{gwtablehead}
    & \multicolumn{3}{cI}{\textbf{With sheet}} & \multicolumn{3}{c}{\textbf{Sheet removed}} \\
    \rowcolor{gwtablehead}
    \textbf{Model} & \textbf{Own $\uparrow$} & \textbf{Others} & \textbf{Margin} & \textbf{Own $\uparrow$} & \textbf{Others} & \textbf{Margin} \\
    \hlineB{1.5}
    BD-Sheet & 0.588 $\pm$ 0.096 & 0.481 & 0.107 & 0.564 $\pm$ 0.102 & 0.463 & 0.101 \\
    \zr DMD-Sheet & 0.584 $\pm$ 0.072 & 0.474 & 0.110 & 0.570 $\pm$ 0.079 & 0.461 & 0.110 \\
    \hlineB{2.5}
  \end{tabular}
\end{table}

\subsection{Distilled Causal Model}
\label{sec:exp-dmd}

\begin{figure}[!tbp]
  \centering
  \includegraphics[width=\textwidth]{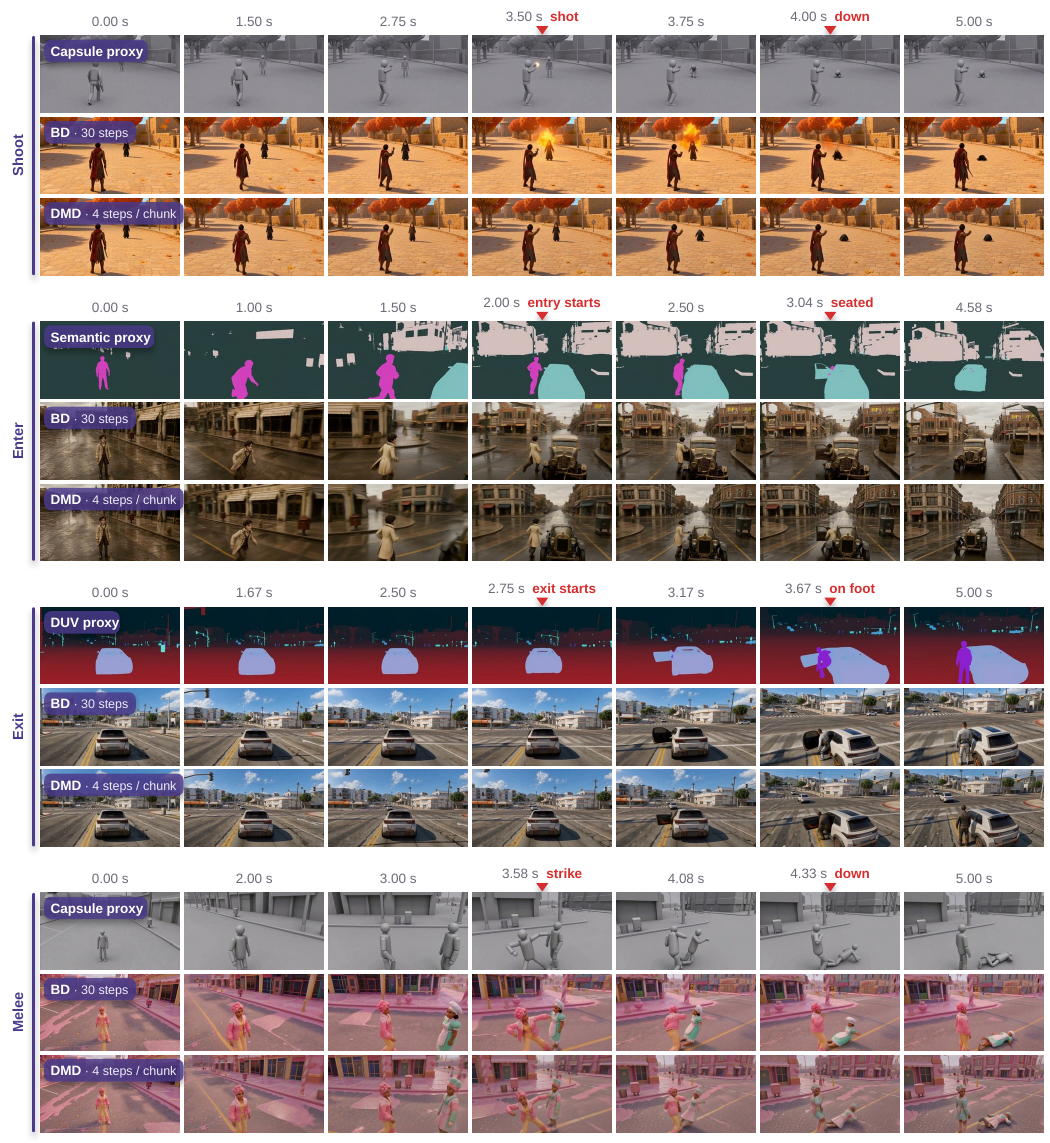}
  \caption{\textbf{Bidirectional teacher and causal student on four events.} Each group shows the proxy above the bidirectional output (30 steps, whole clip) and the DMD output (4 steps per chunk). Shoot: the shooting clip with its capsule proxy; Enter, Exit, Melee: held-out clips under semantic, DUV, and capsule proxies. Columns are chosen per event; red ticks mark the engine's event frames (for the two attacks, read off the proxy).}
  \label{fig:exp-dmd-stages}
\end{figure}

\paragraph{From bidirectional to four-step causal.} Figure~\ref{fig:exp-dmd-stages} compares the bidirectional teacher with the DMD student on the shooting clip and three held-out events. Like the teacher, the student completes every event at the engine's frames and follows the proxy's camera and layout; it stays as sharp as the teacher, and no seam is visible where chunks meet. What weakens is what the proxy does not carry: the fireball is faint under the student, and DMD-Sheet renders fainter skills than BD-Sheet for all eight characters (Figure~\ref{fig:app-oc-skills}). Each character keeps its look through the eight chunks, with identity scores matching BD-Sheet (Table~\ref{tab:exp-identity}, Figure~\ref{fig:app-oc-scenes-dmd}).

\begin{figure}[tp]
  \centering
  \includegraphics[width=\textwidth]{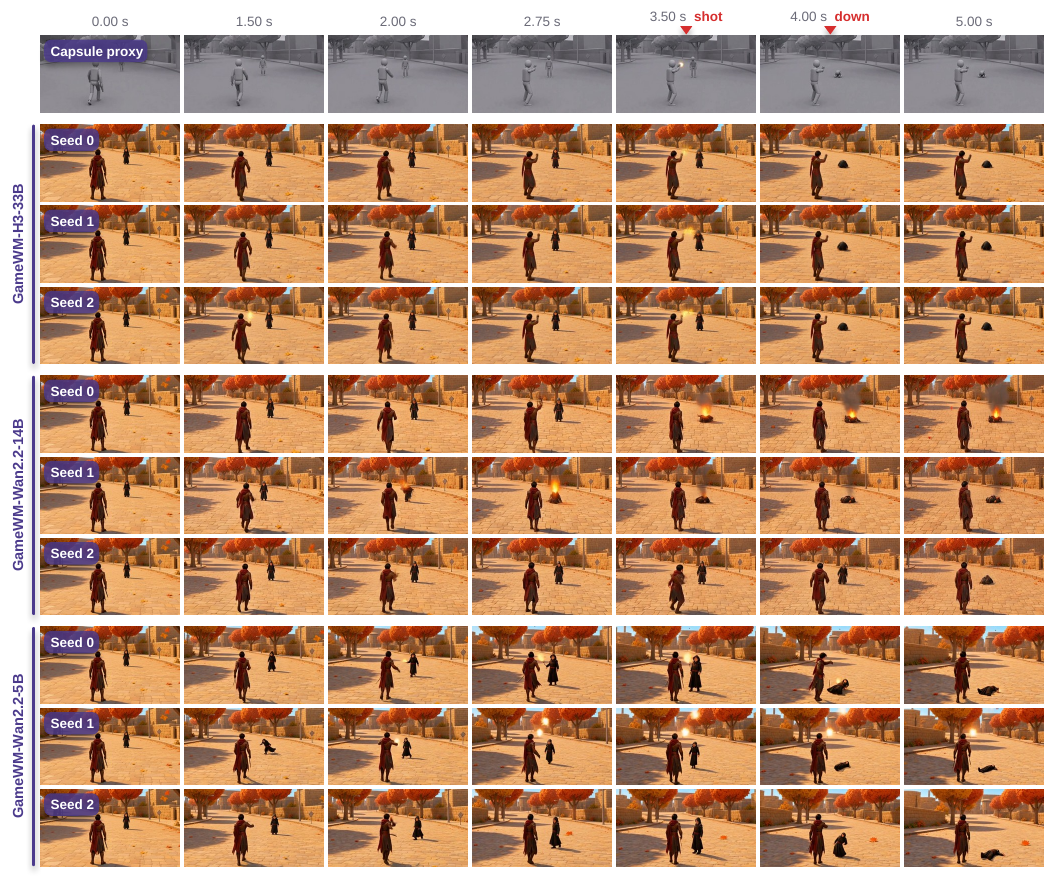}
  \caption{\textbf{Four-step causal students across backbones.} The same capsule proxy (top), fire-mage sheet, first frame, and caption for every model; for each model, the three of eight seeds that follow the proxy most closely. GameWM-H3-33B is DMD-Sheet; the Wan2.2 students are distilled from their sheet-conditioned bidirectional models. Columns are times on the 24\,fps clip (the 14B student renders 16\,fps); red ticks mark the shot and the fall in the proxy.}
  \label{fig:exp-dmd-scale}

  \vspace{1.2em}
  \captionof{table}{\textbf{Latency on the shooting clip} (124 frames at $1280\times704$, second run after warm-up). SP: sequence-parallel degree, the number of GPUs per sample. ``First chunk'' includes filling the key--value cache; ``Rollout'' ends when the last chunk's latent is done, before VAE decoding; ``Task'' is the whole job, including decoding and writing the video.}
  \label{tab:exp-speed}
  \gwtablestyle
  \begin{tabular}{L{3.0cm}IC{1.9cm}C{2.2cm}C{1.8cm}C{1.4cm}}
    \hlineB{2.5}
    \rowcolor{gwtablehead}
    \textbf{Model} & \textbf{First chunk} & \textbf{Per chunk (17 frames)} & \textbf{Rollout} & \textbf{Task} \\
    \hlineB{1.5}
    BD, 30 steps, SP8 & \multicolumn{3}{c}{\textit{whole clip in one pass}} & 95\,s \\
    \zr BD, 30 steps, SP1 & \multicolumn{3}{c}{\textit{whole clip in one pass}} & ${\approx}$576\,s \\
    \hr \textbf{DMD, 4 steps, SP8} & 3.1\,s & 0.94--1.30\,s & 11.5\,s & 28.2\,s \\
    \zr DMD, 4 steps, SP1 & 10.1\,s & 4.8--7.8\,s & 58.4\,s & 78\,s \\
    \hlineB{2.5}
  \end{tabular}
\end{figure}

\paragraph{Across backbones.} Figure~\ref{fig:exp-dmd-scale} runs the four-step students of all three backbones (Table~\ref{tab:models}) on the fire mage's shot, with eight seeds each. Only the 33B student keeps the event on the proxy's clock in all eight. The 14B student renders the strongest fire but hits the opponent a second or more early in five of eight seeds; the 5B student drops the opponent near the proxy's down frame but changes the staging: the opponent walks up, a second figure falls, or the camera closes in. Appearance survives distillation at every scale; what the smaller students lose is the timing and staging they must infer from a capsule's motion.

\paragraph{Speed.} Table~\ref{tab:exp-speed} times both models on one node of eight GPUs. The bidirectional model shows nothing until the whole clip is done (95\,s). The causal model shows its first chunk after 3.1\,s, of which 2.2\,s fill the key--value cache, and then generates 13--18 frames per second, still below the 24\,fps playback rate; VAE decoding still runs once over the whole clip, so chunks are not yet displayed as they are generated. Matrix-Game 4.0 analyzes real-time deployment of a 5B proxy-conditioned renderer in detail~\citep{matrixgame2026v4}.

The proxy-position ablation and a summary of scaling are in Appendix~\ref{sec:app-quant}.

%% file: sec/9_conclusion.tex
\section{Conclusion}
\label{sec:conclusion}
\label{sec:limitations}

GameWM makes a video model a component of a game rather than its replacement. The GameWM-engine keeps the world and computes every event, and it hands each event to the GameWM-VM as a coarse proxy video and an event log; the GameWM-VM decides how the event plays out, as the character and the skills authored for it. An interaction-centric data engine builds training pairs for this contract from game recordings, screen recordings, and simulator footage, and one interface carries it across backbones from 5B to 33B parameters. On held-out clips, the models complete each interaction on the engine's schedule from proxies as coarse as capsules, realize one mechanic as the skills of many characters, and keep the timing after distillation into four-step causal students.

%% file: sec/A_appendix.tex
\section{Data Engine Details}
\label{sec:app-data}

\paragraph{Proxy format.}
\label{sec:app-proxy}
All sources are written in one format, whether depth and semantics come from engine buffers or from monocular estimates. Semantics use the twelve classes of CWM's released DUV format~\citep{chen2026cwm}; engine semantics arrive with the recording's own classes (five in the main GTA~V recordings: static world, sky, player, pedestrians, and vehicles) and are mapped onto them. In DUV, the R channel stores inverse log depth over 0.3--256\,m and the G and B channels store a lattice code of the semantic class. DUV is encoded losslessly and verified bit by bit on read-back. Each low-poly style starts from a first frame restyled into that style by an image model and is rendered by H3 Ref2VA from the clip's semantic video and event text, as the high-fidelity targets are (Section~\ref{sec:data}).

\paragraph{Render check.} Each low-poly and high-fidelity render is compared with its source clip by Gemini~\citep{google2026gemini38flash} before it enters training. For game clips, whose proxies are exact, this is the only quality filter. The VLM scores separate axes: camera (trajectory, framing, and view angle), motion (action type, direction, and phase timing), spatial relations (layout and depth ordering of the subjects), and style (adherence to the requested look); renders with a character sheet add identity, scored for every sheet subject, with the minimum over subjects as the axis score, and the current prompt adds scene layout, whether the static environment keeps the source's layout. Each axis returns a score, a confidence, tags such as \emph{viewpoint} or \emph{unverifiable}, and time-stamped evidence that pairs what the source shows with what the render shows. Each axis is scored from 0 to 4, and a render passes only if every axis scores at least 3; a rejected render is queued for human check, and an axis marked unverifiable, for example a subject fully covered by a mask, never counts as passed. Of 1,414 recent verdicts, 72\% passed; in the rejects the failing axis was most often scene layout, followed by spatial relations, motion, camera, and style, and almost never identity. Pass rates per source are reported in Table~\ref{tab:data-snapshot}: exact engine and simulator proxies give the highest rates (GTA~V, and 75\% of the CARLA low-poly renders), estimated depth and semantics (ABot-World) and sparse gray-box geometry (Magpie) about half or less. For GTA~V and ABot-World, high-fidelity renders pass less often than low-poly ones, since they must also hold the style and the identity of the anchor.

\begin{table}[t]
  \centering
  \caption{\textbf{Corpus snapshot (2026-11-01).} Source clips in the stylization pool and the share of renders that pass the VLM screen, before human review. The corpus keeps growing as sources are converted; Section~\ref{sec:exp-setup} states which data each checkpoint was trained on.}
  \label{tab:data-snapshot}
  \gwtablestyle
  \begin{tabularx}{\textwidth}{L{2.0cm}YL{4.0cm}rrr}
    \hlineB{2.5}
    \rowcolor{gwtablehead}
    \textbf{Source} & \textbf{Proxy obtained from} & \textbf{Engine events} & \textbf{Clips} & \textbf{Low-poly pass} & \textbf{High-fid.\ pass} \\
    \hlineB{1.5}
    GTA~V & Engine buffers (exact) & Fine-grained and coarse & 1{,}460 & 94\% & 70\% \\
    \zr ABot-World & Monocular estimates & Coarse (exploration) & 2{,}765 & 55\% & 49\% \\
    CARLA & Simulator (exact) & None (street traffic) & 1{,}000 & 75\% & -- \\
    \zr Magpie & Simulator, gray-box geometry & Coarse (locomotion) & 1{,}030 & 30\% & 49\% \\
    \hlineB{2.5}
  \end{tabularx}
\end{table}

\paragraph{Window-scoped captions.} The captioner receives the clip, a grid of time bins, and a per-second evidence table decoded from the DUV proxy without any model call: the number of people and vehicles on screen, whether the protagonist is visible, the depth range, and whether anything moved. Its reply is a small JSON document of observations; the caption text, its ordering, and the timestamps are rendered by code. The captioner selects bin indices rather than writing clock times, and names the protagonist's facing in screen space from a closed list, since free-form directions such as ``facing forward'' were read in opposite ways by writer and reader. A malformed reply is returned to the captioner with the list of problems for one retry.

\paragraph{Caption verification.} Each caption is then held against the same evidence table. A claim the proxy contradicts outright, such as a vehicle in a second with no vehicle pixels or a protagonist in a second with no protagonist pixels, is a \emph{fail}; a disagreement with an innocent explanation, such as a bystander count off by one or a still subject while a follow camera moves, is a \emph{warn}. Fails mark hallucinations, which would teach the model to generate objects the proxy does not ask for; warns are worth counting across the corpus but not worth dropping a clip for. The two rates are tracked separately: a rising fail rate means the captioner is losing its grip on the evidence, while a rising warn rate usually means a verification threshold needs to move. Verification currently classifies only: fail captions are counted and reported, and their clips are not removed from training.

\paragraph{Proxy quality review.} This review applies to proxies estimated from screen recordings; proxies unpacked from engine buffers are exact and are not reviewed. Estimated proxies can fail in ways that no per-frame check sees: a region whose depth pulses over time, a road labeled as vegetation for a few frames, a person the segmenter never found. Each clip first passes structural audits, cross-clip depth-scale checks, and sky-depth repair. For each clip a $2\times2$ panel is rendered: RGB, depth (near warm, far blue, sky black), semantics in a fixed 12-class palette, and semantics overlaid on RGB. The panel is sampled at 8\,fps, which catches a one-frame flash at 24\,fps one time in three and a defect lasting a few frames almost always, and sent to Gemini~\citep{google2026gemini38flash} with the meaning of each color and the definitions of flicker and error; HUD elements are not scored. The VLM returns four scores from 1 to 5 (depth temporal stability, depth accuracy, semantic temporal stability, semantic accuracy), a list of time-stamped issues, each with its track, kind (flicker, scale jump, wrong label, missing object, boundary, hole, other) and severity (minor or major), and its own verdict. A clip is rejected if the VLM rejects it, if any score is below 3, or if any issue is major; the clip score is the mean of the four scores divided by 5. The thresholds and the prompt version are stored with each decision.

\paragraph{Local flicker measures.} Two measures are computed without a model so that the VLM can be checked against something it cannot argue with. Depth jitter is the median over pixels of the second temporal difference $|c_{t+1} - 2c_t + c_{t-1}|$ of the log-depth code, so that smooth motion cancels and only jitter remains. Semantic flicker is the share of pixels whose label at $t$ differs from the labels at $t\pm3$ while those two agree, i.e.\ a flash of at most five frames; moving edges under monotonic motion do not count. A span of one frame is not enough, because SAM~2 propagation already removes one-frame flashes. These measures are used for sorting and cross-checking and do not enter the decision.

\paragraph{Human review.} A review workbench lists accepted, rejected, and pending clips with their scores, issues, and measures, plays the review panel frame by frame, and records overrides with reviewer and time. A human verdict takes precedence over the VLM, and dataset splits follow the final verdict.

\section{Evaluation Protocol}
\label{sec:app-protocol}
\label{sec:eval-protocol}

No public benchmark yet exists for video generation referenced by a proxy video. General benchmarks score visual quality or text alignment~\citep{huang2024vbench}, and world-model benchmarks score action-conditioned navigation, memory, and multi-turn interaction~\citep{ye2026mind,ying2026wbench,lu2026wrbench}; neither checks whether a generated clip follows a proxy frame by frame or completes the interaction the engine specified. Proxy-conditioned systems also differ in proxy format, resolution, and conditioning interface, so numbers measured on one system's proxies do not transfer to another. We therefore compare each output directly with its proxy on held-out clips and compare only our own models and checkpoints with each other.

\paragraph{Held-out clips.} Clips are split by source recording, so no recording contributes to both training and evaluation, and stratified by target type, proxy modality, and action. Subsets cover clips with engine-buffer proxies, screen recordings with estimated proxies, and fast camera motion.

\paragraph{Visual comparison.} The report relies on visual comparison rather than aggregate metrics. Every figure places the proxy above the output it drives on shared time columns and marks the engine's event frames, so that adherence to layout and camera, the completion of an interaction, and its timing are judged frame by frame against the engine rather than against another model. Event times come from the engine record or, where none was kept, are read off the proxy; the frames at which the generated interaction happens are read off the output. Two quantities are measured where a visual judgment would be ambiguous: protagonist identity, as DINOv2~\citep{oquab2023dinov2} similarity between the tracked protagonist and its character sheet (Table~\ref{tab:exp-identity}), and latency (Table~\ref{tab:exp-speed}).

\paragraph{Identity metric.} People are detected with DETR~\citep{carion2020detr} every six frames, and the protagonist is tracked by box overlap from its box in the first proxy frame; a track stops once the protagonist enters a vehicle. The score is the cosine similarity between the DINOv2 CLS feature of each crop and the three views of the sheet, maximized over views since the protagonist mostly faces away, and averaged over the track. The same quantity averaged over the other three characters' sheets is the cross-identity baseline.

\paragraph{Comparisons.} External systems are not shown: most do not release weights, and their text prompts cannot be aligned fairly with ours. The ablation (Appendix~\ref{sec:eval-ablations}) compares two of our own settings on the same held-out clips.

\section{Ablations and Scaling}
\label{sec:app-quant}

\subsection{Ablations}
\label{sec:eval-ablations}

\paragraph{Proxy position.} The proxy can enter the sequence as a prefix with its own temporal coordinates, or aligned in time, so that each proxy frame takes the coordinate of the target frame it describes. In an early round on a smaller GTA~V corpus the two settings reached the same reconstruction quality on held-out clips after 1,000 steps (PSNR 18.57 and 18.54\,dB, SSIM 0.598 for both). With no measurable difference, all later checkpoints, including every model in this report, use the sequential prefix.

\subsection{Scaling from 5B to 33B}
\label{sec:eval-scaling}
The three backbones of Table~\ref{tab:models} are trained on the same corpus with the same interface and distilled by the same three stages. Figure~\ref{fig:exp-dmd-scale} compares their four-step students on the shooting clip: the scene and the character survive at every scale, while the 14B student fires a second or more early in five of eight seeds and the 5B student loses the staging of the opponent.

\section{Additional Visual Results}
\label{sec:app-visual}

\begin{figure}[!p]
  \centering
  \includegraphics[width=\textwidth]{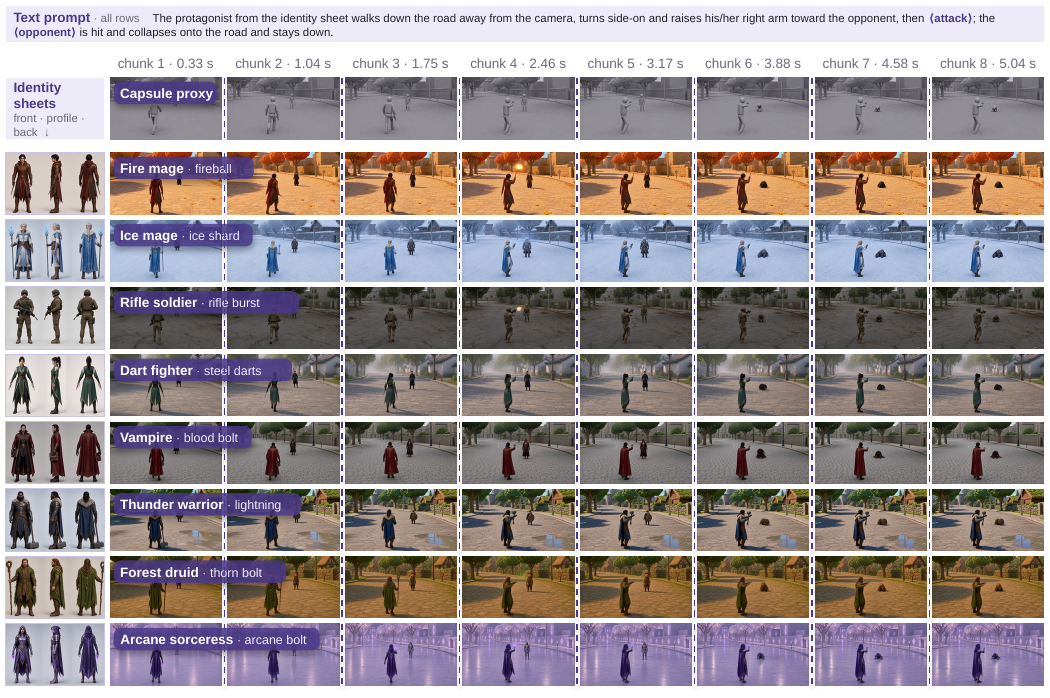}
  \caption{\textbf{Characters and their skills, causal model.} One capsule proxy and one caption template (top); each row adds a character sheet (left), a first frame of that character, and the character's skill (row label). DMD-Sheet, 4 steps per chunk; one frame from the middle of each chunk, dashed lines at chunk boundaries.}
  \label{fig:app-oc-skills}

  \vspace{14pt}
  \includegraphics[width=\textwidth]{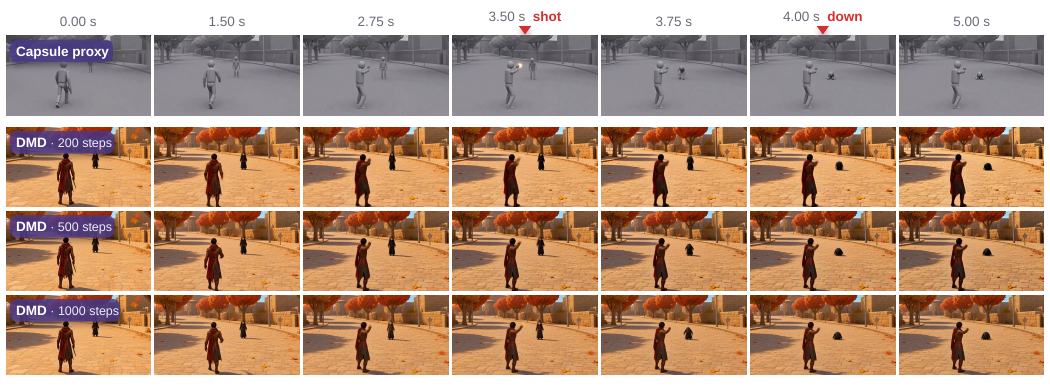}
  \caption{\textbf{DMD outputs over training.} The shooting clip rendered by the causal student after 200, 500, and 1,000 DMD steps, on the columns of the shooting group of Figure~\ref{fig:exp-dmd-stages}; red ticks mark the shot and the fall in the proxy.}
  \label{fig:app-dmd-steps}
\end{figure}

\begin{figure}[!p]
  \centering
  \includegraphics[width=\textwidth,height=0.87\textheight,keepaspectratio]{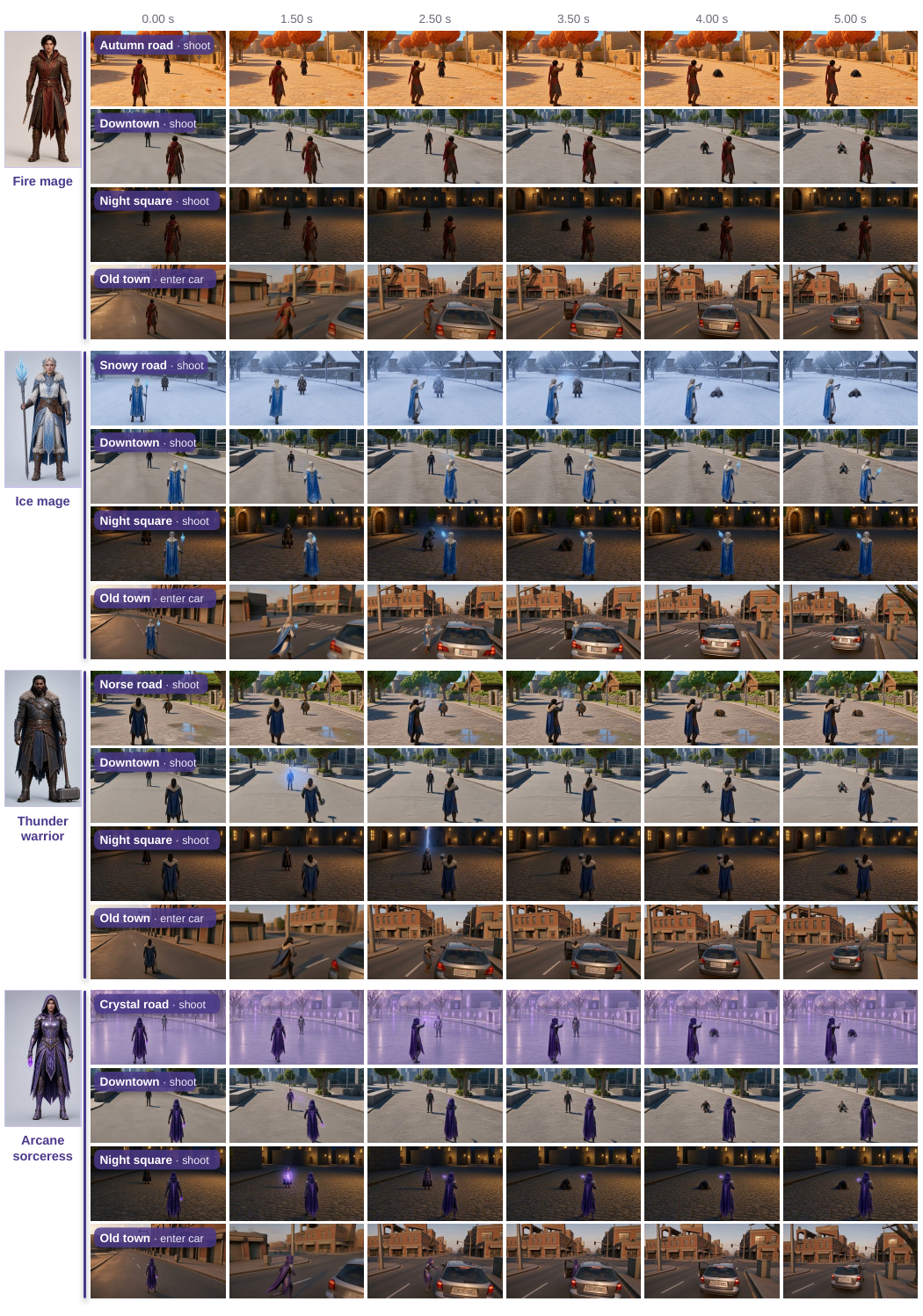}
  \caption{\textbf{One character across scenes, causal model.} The scenes, sheets, first frames, and captions of Figure~\ref{fig:exp-oc-scenes}, rendered by DMD-Sheet at 4 steps per chunk; columns from 0 to 5\,s.}
  \label{fig:app-oc-scenes-dmd}
\end{figure}

\paragraph{Character sheets under causal generation.} Figures~\ref{fig:app-oc-skills} and~\ref{fig:app-oc-scenes-dmd} are the DMD-Sheet versions of the ranged clip in Figure~\ref{fig:exp-oc-mechanics} and of Figure~\ref{fig:exp-oc-scenes}, discussed in Section~\ref{sec:exp-dmd}.

\paragraph{DMD training progression.} In Figure~\ref{fig:app-dmd-steps} the layout, the turn, and the fall are fixed by step 200, with the opponent down by the proxy's frame in all three, and later steps change little that is visible at this scale.

\paragraph{Failure cases.} Figure~\ref{fig:exp-dmd-scale} shows two failures of the smaller causal students on a capsule proxy: a hit that lands a second or more early (14B, five of eight seeds) and an opponent that walks to the protagonist or is joined by a second figure (5B). Four further failures recur in the figures of Section~\ref{sec:exp}. Effects that the proxy does not carry can vanish: in Figure~\ref{fig:exp-oc-mechanics} the rifle butt and the thorn staff are hard to make out, the steel darts are barely visible at this scale, and the fireball is faint under the four-step student (Figure~\ref{fig:exp-dmd-stages}). Identity weakens under side views and occlusion: on the vehicle entry the protagonist's DINOv2 score falls to 0.34--0.52, against 0.49--0.72 on the shooting clips (Table~\ref{tab:exp-identity}). What the proxy omits is filled in by the model rather than by the engine: with the first proxy frame held for the whole clip, the car arrives where the model chooses (Figure~\ref{fig:exp-granularity}). The data engine fails in the same way: the pool's capsule render of a swimming clip followed the strokes but kept the camera pitched down at the water after the recorded camera rose to the horizon, and the render check rejected it on the camera axis.